%% file: main.tex
\documentclass[10pt,twocolumn,letterpaper]{article}

\usepackage{iccv}              

\definecolor{iccvblue}{rgb}{0.21,0.49,0.74}
\usepackage[pagebackref,breaklinks,colorlinks,allcolors=iccvblue]{hyperref}
\usepackage[table,xcdraw]{xcolor}
\usepackage{multirow}
\usepackage[accsupp]{axessibility}  

\newcommand\blfootnote[1]{%
  \begingroup
  \renewcommand\thefootnote{}\footnote{#1}%
  \addtocounter{footnote}{-1}%
  \endgroup
}
\def\paperID{8784} 
\def\confName{ICCV}
\def\confYear{2025}

\title{\hspace{-5pt}\includegraphics[height=1.5em]{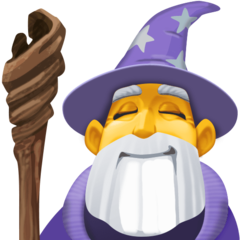}
MagicMotion: Controllable Video Generation with Dense-to-Sparse Trajectory Guidance}

\author{Quanhao Li$^{1,2*}$\qquad Zhen Xing$^{1,2*}$ \qquad Rui Wang$^{1,2}$ \qquad  Hui Zhang$^{1,2}$\qquad  Qi Dai$^{3}$\qquad  Zuxuan Wu$^{1,2\dag}$\\
\normalsize$^1$Shanghai Key Lab of Intell. Info. Processing, School of CS, Fudan University \\
\normalsize$^2$Shanghai Collaborative Innovation Center of Intelligent Visual Computing \\
\qquad \normalsize$^3$ Microsoft Research Asia
\\
\url{https://quanhaol.github.io/magicmotion-site/}
}
\begin{document}
\twocolumn[{
\maketitle
\vspace{-3.4em}
\renewcommand\twocolumn[1][]{#1}
\begin{center}
    \centering
    \includegraphics[width=0.95\textwidth]{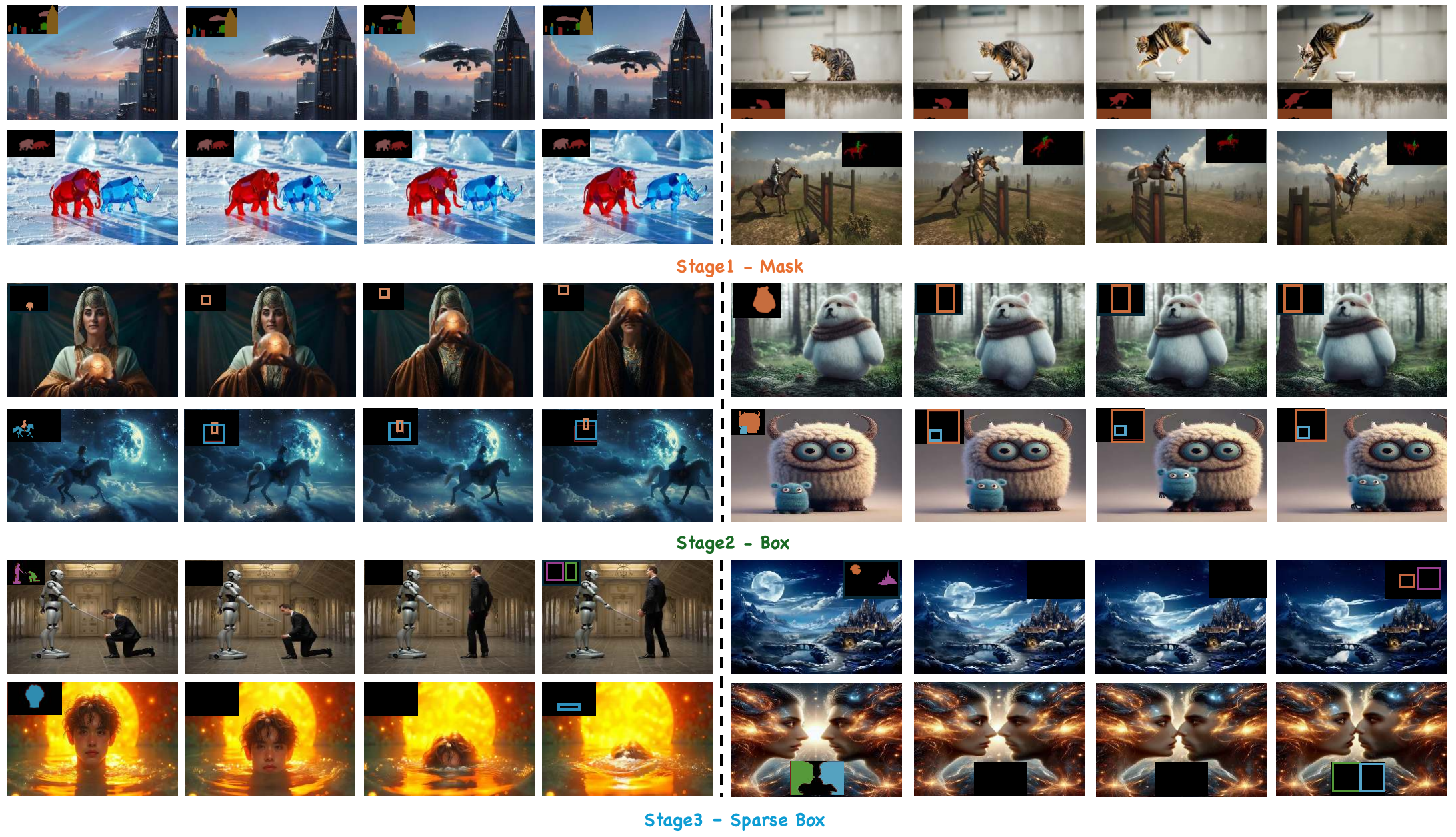}
    \vspace{-0.8em}
    \captionof{figure}{\textbf{Example videos generated by MagicMotion.} MagicMotion consists of three stages, each supporting a different level of control from dense to sparse: mask, box, and sparse box. Given an input image and any form of trajectory, MagicMotion can generate high-quality videos, animating objects in the image to move along the user-specified path.}
    \label{fig:teaser}
\end{center}
}
]
\blfootnote{* equal contributions. $\dag$ Corresponding author.}

\input{sec/0_abstract}

\input{sec/1_intro}
\input{sec/2_relate}
\input{sec/3_method}
\input{sec/4_experiment}
\input{sec/5_conclusion}
{
    \small
    \bibliographystyle{ieeenat_fullname}
    \bibliography{main}
}
\input{sec/X_supplementary}

\end{document}

%% file: sec/0_abstract.tex
\begin{abstract}
Recent advances in video generation have led to remarkable improvements in visual quality and temporal coherence. Upon this, trajectory-controllable video generation has emerged to enable precise object motion control through explicitly defined spatial paths.
However, existing methods struggle with complex object movements and multi-object motion control, resulting in imprecise trajectory adherence, poor object consistency, and compromised visual quality.
Furthermore, these methods only support trajectory control in a single format, limiting their applicability in diverse scenarios.
Additionally, there is no publicly available dataset or benchmark specifically tailored for trajectory-controllable video generation, hindering robust training and systematic evaluation.
To address these challenges, we introduce \textbf{MagicMotion}, a novel image-to-video generation framework that enables trajectory control through three levels of conditions from dense to sparse: masks, bounding boxes, and sparse boxes. Given an input image and trajectories, MagicMotion seamlessly animates objects along defined trajectories while maintaining object consistency and visual quality.
Furthermore, we present \textbf{MagicData}, a large-scale trajectory-controlled video dataset, along with an automated pipeline for annotation and filtering. 
We also introduce \textbf{MagicBench}, a comprehensive benchmark that assesses both video quality and trajectory control accuracy across different numbers of objects.
Extensive experiments demonstrate that MagicMotion outperforms previous methods across various metrics.
\end{abstract}
\vspace{-0.2cm}

%% file: sec/1_intro.tex
\section{Introduction}
\label{sec:intro}
With the rapid development of diffusion models, video generation has witnessed significant progress in recent years. 
In contrast to images, videos~\cite{xing2023svformer, tu2023implicit, wang2022bevt, wang2023masked} capture dynamic temporal changes and offer a richer representation of semantic content.
Earlier video generation approaches, such as AnimateDiff~\cite{guo2023animatediff} and SVD~\cite{blattmann2023stable}, primarily rely on the UNet~\cite{ronneberger2015u} structure, which results in videos with limited length and quality. 
Sora~\cite{videoworldsimulators2024} has demonstrated the powerful capabilities of the DiT~\cite{Peebles2022DiT} architecture in text-to-video (T2V) generation. 
Following this breakthrough, subsequent models leveraging the DiT architecture~\cite{zheng2024open,yang2024cogvideox,kong2024hunyuanvideo} have achieved higher quality outputs and extended video duration.

While DiT-based models~\cite{yang2024cogvideox,kong2024hunyuanvideo, zheng2024open} excel at producing high-quality and longer videos, many text-to-video approaches~\cite{guo2023animatediff, xing2024simda, xing2024survey} lack precise control over attributes like object movement and camera motion~\cite{guo2023animatediff, xing2024simda, yang2024cogvideox}.  Fine-grained trajectory-controllable video generation emerges as a solution, which is especially critical for generating controllable videos in real-world scenarios.

Previous trajectory-controllable video generation methods can be categorized based on the type of control signals they use. 
These include points-based control~\cite{wang2024motionctrl,wang2025levitor,geng2024motionprompting,zhou2025trackgo}, 
optical flow-based control~\cite{zhang2025tora, burgert2025go, yin2023dragnuwa, li2025image, shi2024motion}, 
bounding boxes-based control~\cite{wang2024boximator, wei2024dreamvideo,yang2024direct, qiu2024freetraj, wu2024motionbooth, jain2023peekaboo,namekata2024sgi2v}, 
masks-based control~\cite{yariv2025through}, 
and 3D trajectories-based control~\cite{fu20243dtrajmaster, gu2025das,wang2025cinemaster}.
However, these methods exhibit several limitations. 
First, the trajectory control conditions used by these methods are unitary. Each method only support a single type of control signal.
However, sparse trajectories (\emph{e.g.}, points and optical flow) result in imprecise control over object shape and size, 
while dense trajectories (\emph{e.g.}, masks and 3D trajectories) are challenging for users to provide.
Second, a publicly available large-scale dataset for trajectory-controllable video generation is lacking. 
Existing VOS (Video Object Segmentation) datasets suffer from short video lengths~\cite{xu2018youtube, miao2021vspw, miao2022large}, small scales~\cite{Perazzi2016, MeViS, MOSE, Hong_2023_ICCV}, or few foreground objects~\cite{fan2021lasot}. 
Third, a unified benchmark for evaluating different methods is absent. Besides, previous work focuses solely on video quality and trajectory accuracy while neglecting the impact of the number of moving objects. We argue that controlling fewer or more objects presents different challenges, making it essential to include this factor in the evaluation metrics.

To address these issues, we propose MagicMotion, a controllable video generation  framework with dense-to-sparse trajectory guidance. To inject trajectory conditions into the generation process, we utilize an architecture similar to ControlNet~\cite{zhang2023adding} called \textbf{Trajectory ControlNet} to encode the trajectory information, which is later added to the original DiT model through a zero-initialized convolution layer.
MagicMotion supports three types of trajectory conditions: masks, boxes, and sparse boxes using a progressive training strategy. Experiments show that the model can leverage the knowledge learned in the previous stage to achieve better performance than training from scratch. 
Additionally, we propose a novel latent segment loss that helps the video generation model better understand the fine-grained shape of objects with minimal computation. 

We also construct MagicData, a high-quality public dataset comprising 23K video samples, each annotated with a $ \texttt{<video, text, trajectory>}$ triplet. 
We design a data pipeline that uses a large language model~\cite{touvron2023llama} to extract the main moving objects in the video, and Segment Anything Model (SAM2)~\cite{ren2024grounded, ravi2024sam} to annotate the segmentation masks and bounding boxes of these objects. 
Furthermore, we introduce MagicBench, a large-scale comprehensive trajectory-controllable video generation benchmark. It categorizes all videos into 6 classes based on the number of foreground objects and evaluates models separately for each category in terms of both video generation quality and trajectory control accuracy.

In conclusion, the main contributions of our work are summarized as follows:

\begin{itemize}
    \item We present MagicMotion, a trajectory-controllable image-to-video generation model that supports three types of control signals: masks, boxes, and sparse boxes.
    \item We introduce a data curation and filtering mechanism, and construct MagicData, the first public dataset for trajectory-controlled video generation.
    \item We propose MagicBench, a comprehensive benchmark for evaluating trajectory-controllable video generation models for both video quality and trajectory control accuracy across different numbers of controlled objects.
\end{itemize}

%% file: sec/2_relate.tex
\section{Related Works}
\label{sec:formatting}

\paragraph{Video Diffusion Models}
Diffusion models~\cite{ho2020denoising, song2020denoising, song2020score, liu2022flow} have made great progress in image generation~\cite{dhariwal2021diffusion, nichol2021glide, ramesh2021zero, rombach2022high, saharia2022photorealistic, ramesh2022hierarchical, zhang2024creatilayout, zhang2025creatidesign, zhang2023adadiff, zhang2025blockdance, wang2025simplear, zhang2025reasongen}, which has led to the rapid development of video generation~\cite{ho2022video, wu2023tune, khachatryan2023text2video, guo2023animatediff, blattmann2023stable, zheng2024open, hong2022cogvideo, yang2024cogvideox, kong2024hunyuanvideo, xing2024survey, xing2024aid, xing2023vidiff, xie2025human2robot, du2024challenge, tu2024motioneditor, tu2024motionfollower, tu2025stableanimator, tu2025stableanimator++, tian2025unigen}.
VDM~\cite{ho2022video} is the first to apply diffusion models to video generation. 
Early works like AnimateDiff~\cite{guo2023animatediff} and SimDA~\cite{xing2024simda} attempt to insert temporal layers into pretrained T2I model for video generation.
Subsequently, VideoCrafter~\cite{chen2023videocrafter1} and SVD~\cite{blattmann2023stable} use large-scale and high-quality data for training, acheiving better performance.
However, these methods have trouble on generating long videos with high quality, mainly due to the inherent limitations of the UNet architecture.
The emergence of Sora~\cite{videoworldsimulators2024} is a significant success, demonstrating the potential of DiT~\cite{Peebles2022DiT} models to generate high-quality videos with tens of seconds.
Recent video generation methods~\cite{zheng2024open, yang2024cogvideox, kong2024hunyuanvideo} are mainly based on DiT architecture, and have achieved great success in the open-source community.
However, these methods rely solely on text or image guidance for video generation, lacking precise control over object or camera trajectory, which is crucial for high-quality video generation.

\vspace{-1em}
\paragraph{Trajectory Controllable Video Generation}
Trajectory Controllable Video Generation has recently garnered significant attention for its ability to precisely control object and camera trajectories during video synthesis.
Previous methods~\cite{yin2023dragnuwa, zhang2025tora, shi2024motion, li2025image}
integrate optical flow maps into video generation through a trajectory encoder.
Recent works~\cite{wang2024motionctrl, zhou2025trackgo} suggest using point maps as a form of guidance.
MotionCtrl~\cite{wang2024motionctrl} processes point maps with a Gaussian filter and employs trainable encoders to encode object trajectories. Trackgo~\cite{zhou2025trackgo} represents objects using a few key points and injects this information via an encoder and a custom-designed adapter structure.
Other works~\cite{wang2024boximator,qiu2024freetraj, namekata2024sgi2v, jain2023peekaboo} employ bounding box to control object trajectories. Boximator~\cite{wang2024boximator} employs a trainable self-attention layer to fuse box and visual tokens inspired by GLIGEN~\cite{li2023gligen}. Some training-free methods~\cite{qiu2024freetraj, namekata2024sgi2v, jain2023peekaboo} purpose to modify attention layers or initial noised video latents to inject box signals.
Additionally, certain methods~\cite{wang2025cinemaster, gu2025das, fu20243dtrajmaster,wang2025levitor} explore the potential of 3D trajectories to achieve more sophisticated motion control. LeViTor~\cite{wang2025levitor} employs keypoint trajectory maps enriched with depth information, while others~\cite{wang2025cinemaster, gu2025das, fu20243dtrajmaster} construct custom 3D trajectories to represent object movements.
However, sparse trajectories lead to imprecise control on objects shape and size, while dense trajectories are difficult for users to provide.
In contrast, MagicMotion can control both dense and sparse trajectories, providing users with more flexible control over video generation.

%% file: sec/3_method.tex
\section{Method}

\subsection{Overview}
Our work mainly focuses on trajectory-controllable video generation. Given an input image $I \in R^{H \times W \times 3}$, and several trajectory maps $C \in R^{T \times H \times W \times 3}$, the model can generate a video $V \in R^{T \times H \times W \times 3}$ in line with the provided trajectories, where T denotes the length of generated video.
In the following sections, we first provide a detailed explanation of our model architecture in Section \ref{sec:architecture}. Next, we outline our progressive training procedure in Section \ref{sec:procedure}. In Section \ref{sec:latent segment loss}, we introduce the Latent Segmentation Loss and demonstrate how it enhances the model capabilities on fine-grained object shape. We then describe our dataset curation and filtering pipeline in Section \ref{sec:data pipeline}. Finally, we present an in-depth overview of MagicBench in Section \ref{sec:benchmark}.

\subsection{Model Architecture}
\label{sec:architecture}

\begin{figure*}[ht]
    \centering
    \includegraphics[width=\linewidth]{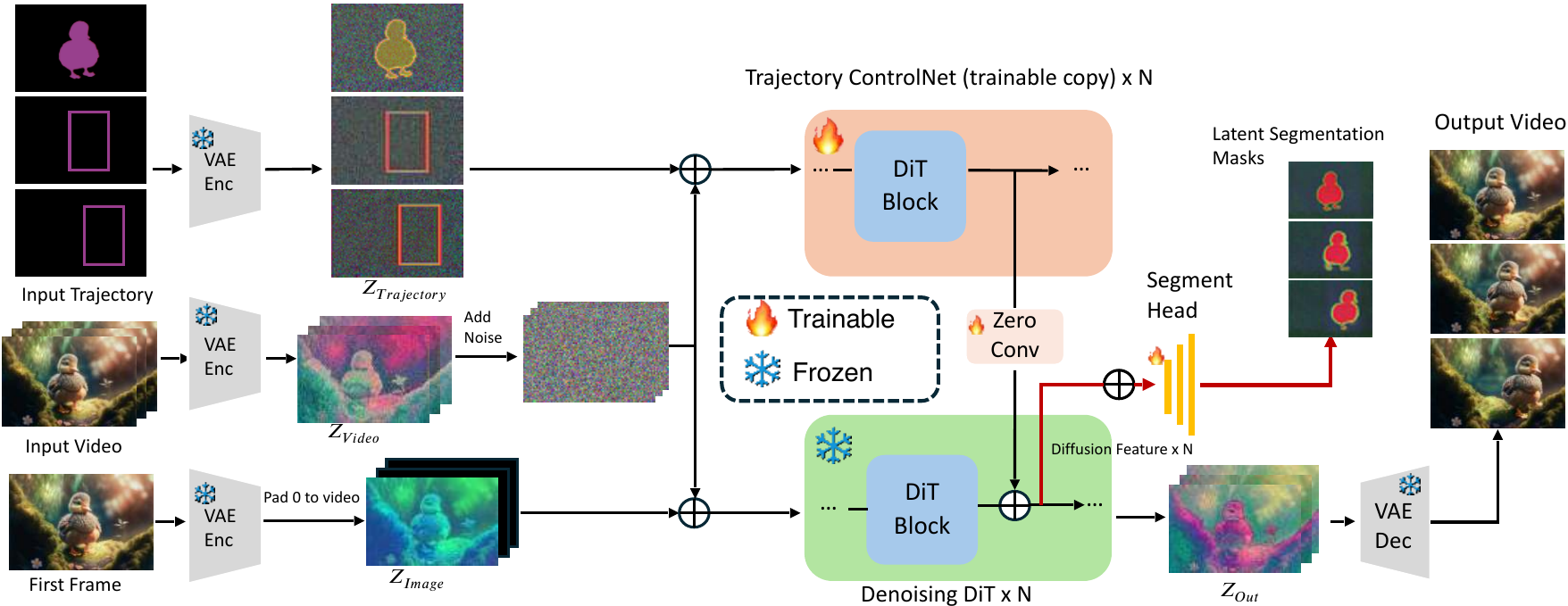}
     \caption{Overview of MagicMotion Architecture (text prompt and encoder are omitted for simplicity). 
     MagicMotion employs a pretrained 3D VAE to encode the input trajectory, first-frame image, and training video into latent space. It has two separate branches: the video branch processes video and image tokens, and the trajectory branch uses Trajectory ControlNet to fuse trajectory and image tokens, which is later integrated to the video branch through a zero-initialized convolution layer.
     Besides, diffusion features from DiT blocks are concatenated and processed by a trainable segment head to predict latent segmentation masks, which contribute to our latent segment loss.
     }
    \label{fig:model_architecture}
\end{figure*}

\paragraph{Base I2V generation model}
We utilize CogVideoX-5B-I2V ~\cite{yang2024cogvideox} and Wan2.1 1.3B~\cite{wan2025} as our base image to video model. The models are built upon a DiT (Diffusion Transformer) architecture, incorporating 3D-Full Attention to generate high-quality videos.
As shown in Fig.~\ref{fig:model_architecture}, the model takes an input image $I \in R^{H \times W \times 3}$ and a corresponding video $V \in R^{T \times H \times W \times 3}$ and encodes them into latent representations $Z_{image},Z_{video} \in R^{\frac{T}{4} \times \frac{H}{8} \times \frac{W}{8} \times 16}$ using a pretrained 3D VAE ~\cite{kingma2013auto}.
Later, $Z_{image}$ is zero-padded to $T$ frames and concatenated with a noised version of $Z_{video}$ and then fed into the Diffusion Transformer, where a series of Transformer blocks iteratively denoise it over a predefined number of steps. 
Finally, the denoised latent is decoded by a 3D VAE decoder to get the output video $V_{out} \in R^{T \times H \times W \times 3}$.

\vspace{0.2cm}
\noindent\textbf{Trajectory ControlNet}
~~To ensure that the generated video follows the motion patterns given by the input trajectory maps $C \in R^{T \times H \times W \times 3}$, we adopt a design similar to ControlNet~\cite{zhang2023adding} to inject trajectory condition. 
As shown in Fig.~\ref{fig:model_architecture}, we employ the 3D VAE encoder to encode the trajectory maps into $Z_{trajectory} \in R^{\frac{T}{4} \times \frac{H}{8} \times \frac{W}{8} \times 16}$, which is then concatenated with the encoded video $Z_{video}$ and serves as input to Trajectory ControlNet.
Specifically, Trajectory ControlNet is constructed with a trainable copy of all pre-trained DiT blocks to encode the user-provided trajectory information. The output of each Trajectory ControlNet block is then processed through a zero-initialized convolution layer and added to the corresponding DiT block in the base model to provide trajectory guidance.

\subsection{Dense-to-Sparse Training Procedure}   
\label{sec:procedure}

Dense trajectory conditions, such as segmentation masks, offer more precise control than sparse conditions like bounding boxes but are less user-friendly. To address this, MagicMotion employs a progressive training procedure, where each stage initializes its model with the weights from the previous stage. This enables three types of trajectory control ranging from dense to sparse. We found that this progressive training strategy helps the model achieve better performance compared to training from scratch with sparse conditions.
Specifically, we adopt the following trajectory conditions across stages: stage1 uses segmentation masks, stage2 uses bounding boxes, and stage3 uses sparse bounding boxes, where fewer than 10 frames have box annotations.
Additionally, we always set the first frame of the trajectory condition as a segmentation mask to specify the foreground objects that should be moving.

Our model uses velocity prediction following~\cite{yang2024cogvideox}. Let $x_0$ be the initial video latents, $\epsilon$ be the gaussian noise, $x_t = \sqrt{\alpha_t} * x_0 + \sqrt{1 - \alpha_t} * \epsilon$ be the noised video latents, and $v_\theta$ be the model output. The diffusion loss can be written as:
\begin{equation}
\resizebox{0.9\hsize}{!}{$
    \mathcal{L}_{diffusion}=
    \mathbb{E}_{t,\epsilon \sim \mathcal{N}(0,I),x_{0}}
    \Big[\left\| x_0 - \Big(\sqrt{\alpha_t}\,x_t - \sqrt{1-\alpha_t}\,v_\theta \Big)\right\|_2^2 \Big]
$}
\end{equation}

\subsection{Latent Segmentation Loss}
\label{sec:latent segment loss}
Bounding box-based trajectory is able to control an object’s position and size but lacks fine-grained shape perception. To address this, we propose Latent Segmentation Loss, which introduces segmentation mask information during model training and enhances the model’s ability to perceive fine-grained object shapes. 

Previous works~\cite{xu2023odise, zhao2023unleashing, bhat2025simgen, weng2024genrec} have leveraged diffusion generation models for perception tasks, demonstrating that the features extracted by diffusion models contain rich semantic information. However, these models generally operate in the pixel space, which leads to extensive computational time and substantial GPU memory.

To incorporate dense trajectory information while keeping computational costs within a reasonable range, we propose utilizing a lightweight segmentation head to predict segmentation masks directly in the latent space, eliminating the need for decoding operations.
Specifically, our segmentation head takes a list of diffusion features $Z_{feature} \in R^{\frac{T}{4} * \frac{H}{16} * \frac{W}{16} * 3072}$ from each DiT block, and outputs a latent segmentation mask $Z_{segment} \in R^{\frac{T}{4} * \frac{H}{8} * \frac{W}{8} * 16}$. We use a light-weight architecture inspired by Panoptic FPN~\cite{kirillov2019panoptic}. Each diffusion feature first passes through a convolution layer to extract visual features $P \in R^{\frac{T}{4} * \frac{H}{16} * \frac{W}{16} * 64}$. The resulting features are then concatenated and processed by another convolution layer followed by an upsampling layer to generate the final latent segmentation mask.
We compute the latent segment loss as the Euclidean distance between $Z_{segment}$ and the ground truth mask trajectory latents $Z_{mask}$, which can be written as,
\begin{equation}
\mathcal L_{seg}=\mathbb{E}_{t,\epsilon \sim \mathcal{N}(0, I),z_{0}}[\left \|  Z_{segment} - Z_{mask}\right \| _{2}^{2} ]
\end{equation}

In practice, $\mathcal L_{seg}$ is only used in stage2 and stage3, providing the model with dense conditions information when trained with sparse conditions. In detail, we set the weight of $\mathcal L_{seg}$ as 0.5, and the original diffusion loss as 1. In total, our final loss function can be written as,
\begin{equation}
    \mathcal L = \mathcal L_{diffusion} + \lambda * \mathcal L_{seg}
\end{equation}
where $\lambda$ is set to 0 in stage1, and 0.5 in stage2 and stage3.

\subsection{Data Pipeline}
\label{sec:data pipeline}

\begin{figure}[ht]
    \centering
    \includegraphics[width=\linewidth]{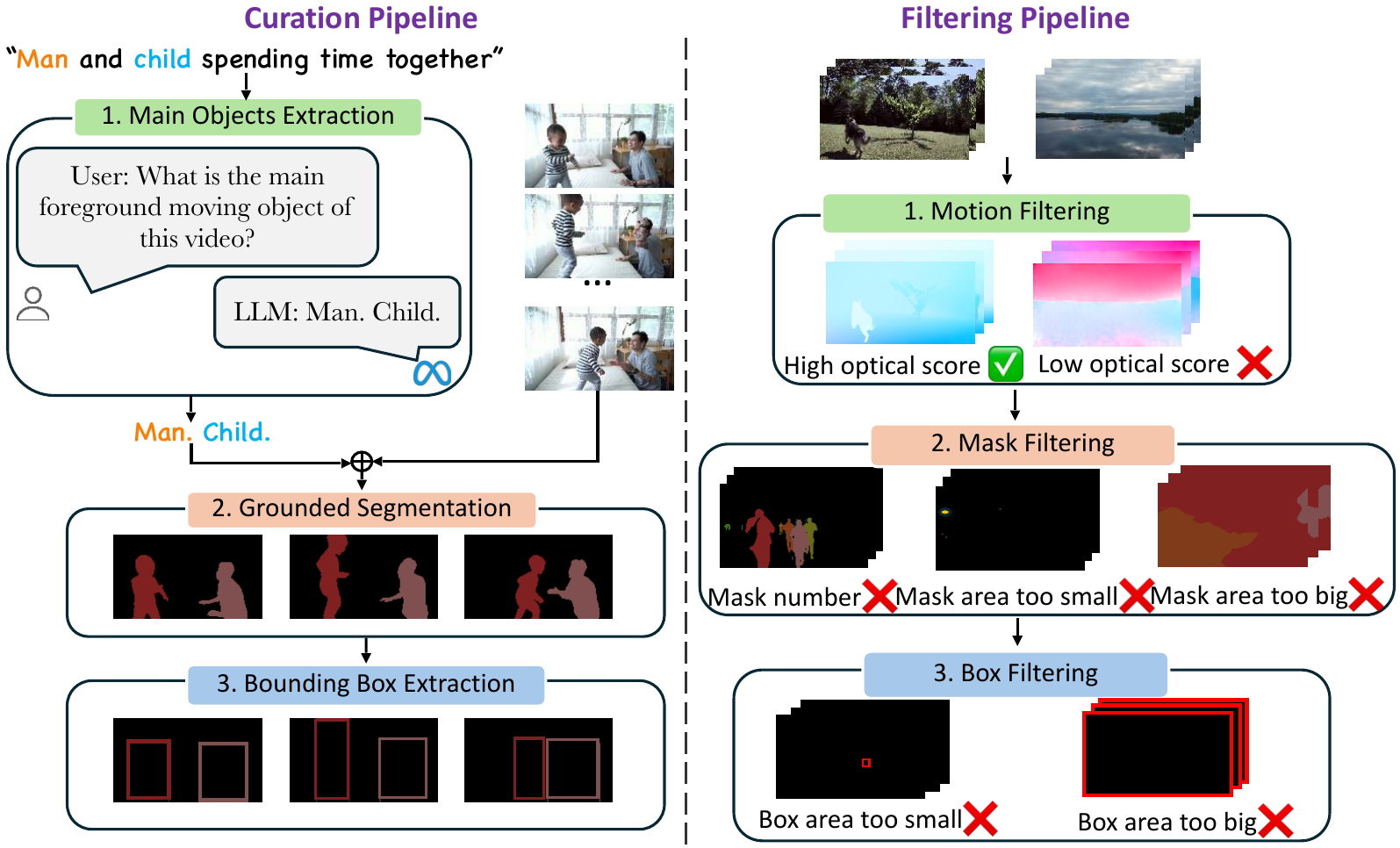}
    \caption{Overview of the Dataset Pipeline. The Curation Pipeline is used to construct trajectory annotations, while the Filtering Pipeline filters out unsuitable videos for training.}
    \label{fig:dataset_pipeline}
\end{figure}

Trajectory controllable video generation requires a video dataset with trajectory annotations. However, existing large-scale video datasets~\cite{bain2021frozen, chen2024panda, ju2024miradata} only provide text annotations and lack trajectory data. Moreover, almost all previous works~\cite{zhang2025tora, zhou2025trackgo, wang2024motionctrl, li2025image, yin2023dragnuwa} use privately curated datasets, which are not publicly available.

We present a comprehensive and general data pipeline for generating high-quality video data with both dense (mask) and sparse (bounding box) annotations.
As shown in Fig.~\ref{fig:dataset_pipeline}, the pipeline consists of two main stages: the Curation Pipeline and the Filtering Pipeline. The Curation Pipeline is responsible for constructing trajectory information from a video-text dataset, while the Filtering Pipeline ensures that unsuitable videos are removed before training.

\vspace{0.2cm}
\noindent\textbf{Curation Pipeline.} 
~~We begin our dataset curation process with Pexels~\cite{Pexels-400k}, a large-scale video-text dataset containing 396K video clips with text annotations. It encompasses videos featuring diverse subjects, various scenes, and a wide range of movements.
We utilize Llama3.1~\cite{touvron2023llama} to extract the foreground moving objects from the textual annotations of each video. As shown in Fig.~\ref{fig:dataset_pipeline}, we input the video’s caption into the language model and prompt it to identify the main foreground objects mentioned in the sentence. If the model determines that the sentence does not contain any foreground objects, it simply returns “empty” and such videos are filtered out.
Next, we utilize Grounded-SAM2~\cite{ravi2024sam, ren2024grounded}, a grounded segmentation model that takes a video along with its main objects as input and generates segmentation masks for each primary object. Each object is consistently annotated with a unique color.
Finally, bounding boxes are extracted from each segmentation mask using the coordinates of the top-left and bottom-right corners to draw the corresponding boxes. The color of the bounding box for each object remains consistent with its masks.

\vspace{0.2cm}
\noindent\textbf{Filtering Pipeline.} 
~~Many videos contain only static scenes, which are not beneficial for training trajectory-controlled video generation models. To address this, we use optical flow scores to filter out videos with little motion and dynamics. Specifically, we utilize UniMatch~\cite{xu2023unifying} to extract optical flow maps between frames and compute the mean absolute value of these flow maps as the optical flow score, representing the video’s motion intensity. However, videos with background movement but static foreground can still have high motion scores. To address this, we further use UniMatch to extract optical flow scores for foreground objects based on segmentation masks and bounding boxes. Videos with low foreground optical flow scores are filtered out, ensuring MagicData includes only videos with moving foreground objects.

The trajectory annotations generated by the curation pipeline require further refinement. As shown in Fig.~\ref{fig:dataset_pipeline}, some videos contain too many foreground objects annotations, or their sizes may be too large or too small. To address this, we regulate these factors within a reasonable range and filter out videos that fall outside the acceptable range.

Specifically, based on extensive manual evaluation, we empirically set the optical flow score threshold to 2.0, limit the number of foreground object annotations from 1 to 3, and constrain the annotated area ratio to a range of 0.008 to 0.83. The whole data curation and filtering pipeline yields us with MagicData, a high quality dataset for trajectory controllable video generation containing 23K videos with both dense and sparse trajectory annotations.

\begin{table*}[ht]
\centering
\resizebox{\linewidth}{!}{
\begin{tabular}{@{}l|cccccccc@{}}
\toprule
\multirow{2}{*}{\textbf{Method}}                                                      & \multicolumn{4}{c}{\textbf{MagicBench}}                                                                  & \multicolumn{4}{c}{\textbf{DAVIS}}                                                  \\ \cmidrule(l){2-9} 
                                                                                      & FID($\downarrow$) & FVD($\downarrow$) & M\_IoU\%($\uparrow$) & \multicolumn{1}{c|}{B\_IoU\%($\uparrow$)} & FID($\downarrow$) & FVD($\downarrow$) & M\_IoU\%($\uparrow$) & B\_IoU\%($\uparrow$) \\ \midrule
Motion-I2V~\cite{shi2024motion}                                 & 39.31             & 354.10            & 56.19                & \multicolumn{1}{c|}{60.66}                & 94.14             & 1558.32           & 32.00                & 42.61                \\
ImageConductor~\cite{li2025image}      & 49.95             & 331.76            & 51.76                & \multicolumn{1}{c|}{52.90}                & 91.71             & 1155.08           & 34.39                & 43.41                \\
DragAnything~\cite{wu2024draganything}                          & 31.36             & 253.40            & 66.30                & \multicolumn{1}{c|}{70.85}                & 70.70             & 1166.22           & 40.13                & 53.60                \\
LeViTor~\cite{wang2025levitor}                                  & 38.32             & 194.53            & 39.96                & \multicolumn{1}{c|}{46.36}                & 97.98             & 922.68            & 25.24                & 31.42                \\
DragNUWA~\cite{yin2023dragnuwa}                                 & 39.73             & 185.52            & 66.88                & \multicolumn{1}{c|}{69.21}                & 84.61             & 1079.89           & 41.22                & 52.61                \\
SG-I2V~\cite{namekata2024sgi2v}                                 & 32.60             & 168.82            & 68.78                & \multicolumn{1}{c|}{74.39}                & 90.93             & 1170.60           & 37.36                & 50.96                \\
Tora~\cite{zhang2025tora} & 26.27             & 245.23            & 58.95                & \multicolumn{1}{c|}{64.03}                & 51.75             & 766.76            & 37.98                & 50.90                \\ \midrule
\textbf{Ours (Stage1-Wan1.3B)}                                                        & \textbf{17.74}         & \textbf{129.79}         & \textbf{83.79}            & \multicolumn{1}{c|}{\textbf{80.96}}            & \textbf{46.65}         & \textbf{548.56}         & \textbf{42.83}           & \textbf{53.67}            \\
\textbf{Ours (Stage2-Wan1.3B)}                                                        & \textbf{19.33}    & \textbf{143.07}   & \textbf{68.89}       & \multicolumn{1}{c|}{\textbf{71.74}}       & \textbf{48.33}    & \textbf{662.82}   & \textbf{44.75}       & \textbf{61.88}       \\ \midrule
\textbf{Ours (Stage1-CogVideoX)}                                                      & \textbf{15.06}    & \textbf{112.69}   & \textbf{91.57}       & \multicolumn{1}{c|}{\textbf{87.75}}       & \textbf{45.06}    & \textbf{579.94}   & \textbf{81.33}       & \textbf{84.97}       \\
\textbf{Ours (Stage2-CogVideoX)}                                                      & \textbf{15.17}    & \textbf{107.21}   & \textbf{76.61}       & \multicolumn{1}{c|}{\textbf{81.45}}       & \textbf{50.36}    & \textbf{760.95}   & \textbf{53.94}       & \textbf{72.84}       \\ \bottomrule
\end{tabular}
}
\caption{Quantitative Comparison results on MagicBench and DAVIS. ``M\_IoU" and ``B\_IoU" refer to Mask\_IoU and Box\_IoU, respectively.}
\label{tab:comparison}
\end{table*}

\begin{figure*}[ht]
    \centering
    \vspace{-0.8em}
    \includegraphics[width=\linewidth]{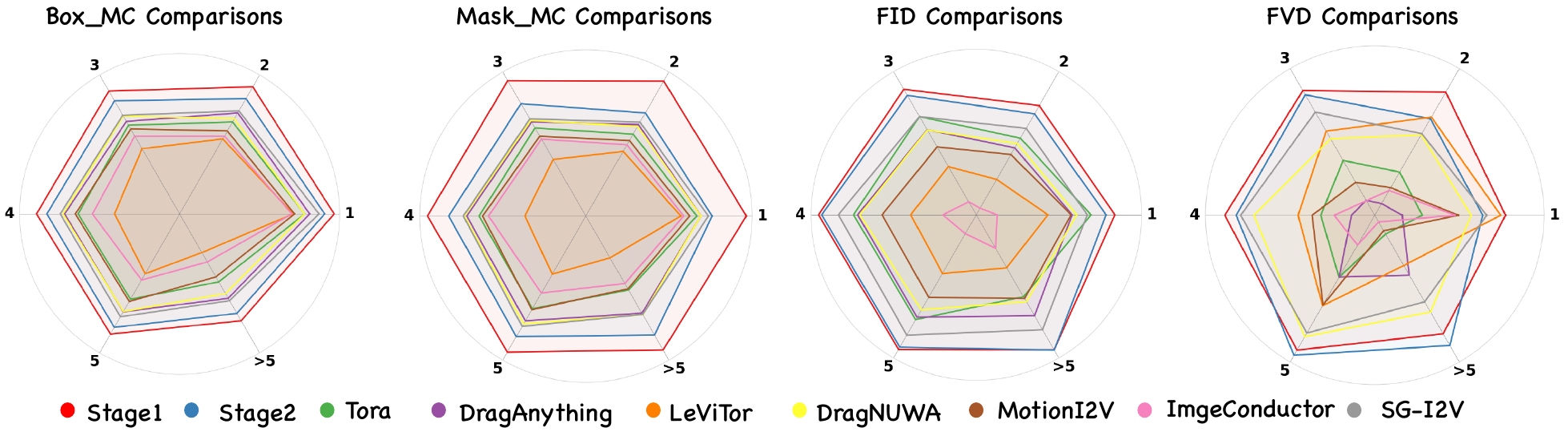}
    \caption{Comparison results of different object number on MagicBench. To present the results more clearly, we have negated the FVD and FID scores.}
    \label{fig:comparisons_radar}
    \vspace{-1.4em}
\end{figure*}

\subsection{MagicBench}
\label{sec:benchmark}
Previous works on trajectory-controlled video generation~\cite{zhang2025tora, li2025image, shi2024motion, objctrl2.5d, zhou2025trackgo, wu2024draganything, namekata2024sgi2v} have primarily been validated on DAVIS (which has a relatively small dataset size), VIPSeg (where the annotated frames per video are insufficient), or privately constructed test sets. Thus there is an urgent need for a large-scale, publicly available benchmark to enable fair comparisons across different models in this field.
To merge this gap, we use the data pipeline mentioned in Sec.~\ref{sec:data pipeline} to construct MagicBench, a large-scale open benchmark consisting of 600 videos with corresponding trajectory annotations. MagicBench evaluates not only video quality and trajectory accuracy but also takes the number of controlled objects as a key evaluation factor. Specifically, it is categorized into 6 groups based on the number of controlled objects, ranging from 1 to 5 objects and more than 5 objects, with each category containing 100 high-quality videos.

\vspace{0.2cm}
\noindent\textbf{Metrics.}
~~For evaluation metrics, we adopt FVD~\cite{ranftl2020towards} to assess video quality and FID~\cite{heusel2017gans} to evaluate image quality, following~\cite{zhou2025trackgo, wang2024motionctrl, wang2025levitor, wu2024draganything}. To quantify motion control accuracy, we use Mask\_IoU and Box\_IoU, which measure the accuracy of masks and bounding boxes, respectively.
Specifically, given a generated video $V_{gen}$, we use the groundtruth masks of the first frame $M_{gt}(0)$ as input to SAM2~\cite{ravi2024sam} to predict the masks $M_{gen}$ of the foreground objects in $V_{gen}$. For each foreground object, we compute the Intersection over Union (IoU) between $M_{gen}(i)$ and groundtruth masks $M_{gt}(i)$ in each frame, then average these values to obtain Mask\_IoU. Similarly, we compute the IoU between the predicted and groundtruth bounding boxes for each foreground object in each frame and take the average as Box\_IoU.

%% file: sec/4_experiment.tex
\section{Experiment}

\subsection{Experiment Settings}
\paragraph{Implementation details.}
We employ CogVideoX 5B~\cite{yang2024cogvideox} and Wan2.1 1.3B~\cite{wan2025} as our base image-to-video model, which is trained to generate video based on an input image.  
Each stage of MagicMotion was trained on MagicData for one epoch.
The training process consists of three stages. stage1 trains Trajectory ControlNet from scratch. In stage2, Trajectory ControlNet is further refined using the weights from stage1, while Segment Head is trained from scratch. Finally, in stage3, both Trajectory ControlNet and Segment Head continue training initialized with the weights from stage2.
All training experiments were conducted on 4 NVIDIA A100-80G GPUs. We employed AdamW ~\cite{loshchilov2017decoupled} as the optimizer, training with a learning rate of $1e-5$ and a batch size of 1 on each GPU.
During inference, we set steps to 50, the guidance scale to 6, and the weight of Trajectory ControlNet to 1.0 by default.

\vspace{-0.5cm}

\paragraph{Datasets.}
During training, we use MagicData as our training set. MagicData is annotated with dense to sparse trajectory information using the data pipeline described in Section~\ref{sec:data pipeline}. It comprises a total of 23K video samples, annotated with text and trajectory.
For evaluation, we evaluate all methods on both MagicBench and DAVIS~\cite{Perazzi2016}, using the comparison metrics illustrated in Sec~\ref{sec:benchmark}.

\begin{figure*}[h]
    \centering
    \includegraphics[width=\linewidth]{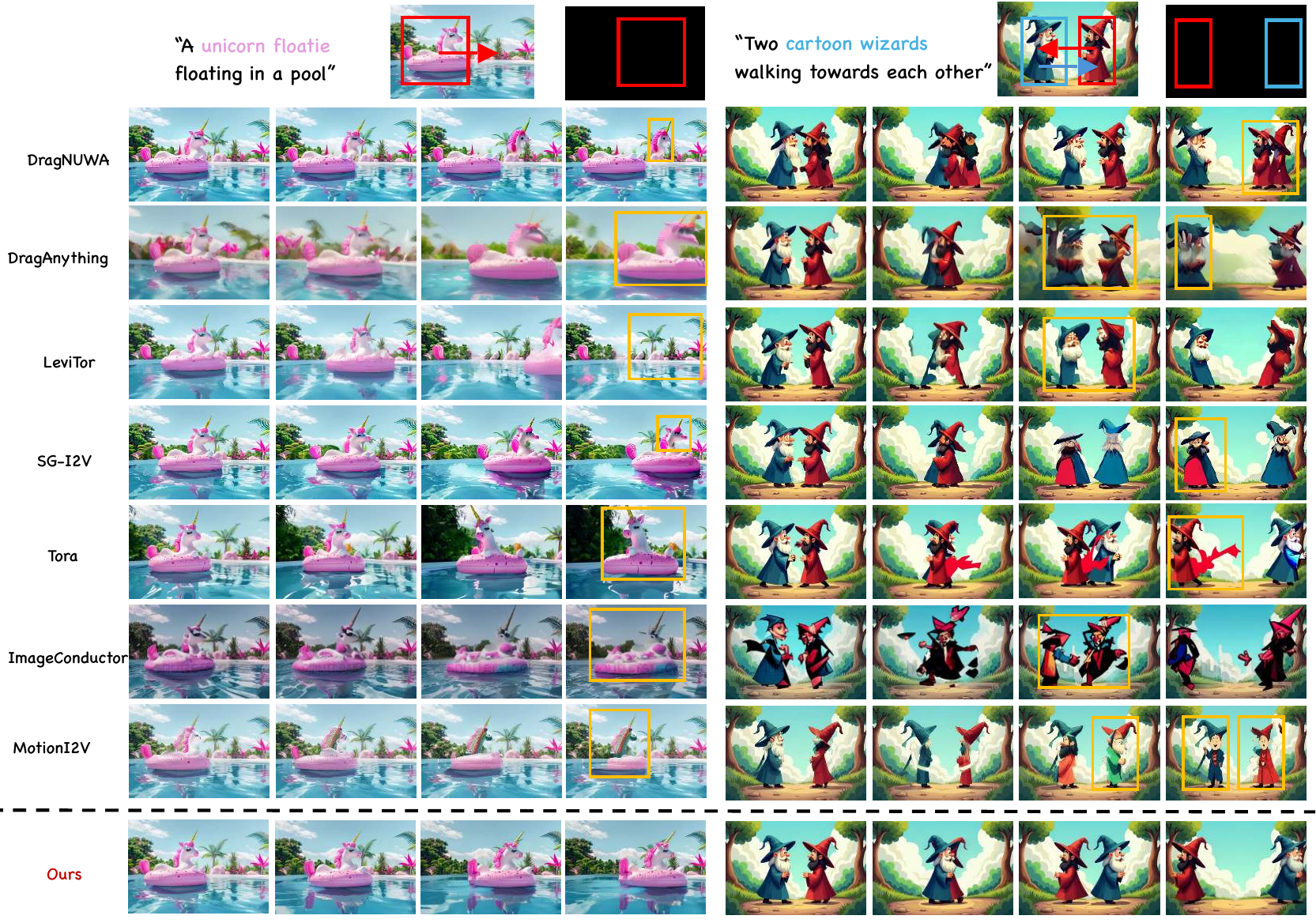}
    \caption{MagicMotion successfully controls the main objects moving along the provided trajectory, while all other methods exhibit significant defects marked with the orange box.}
    \label{fig:comparisons}
\end{figure*}

\subsection{Comparison with Other Approaches}
For thorough and fair comparisons, we compare our methods against 7 public trajectory controllable I2V methods~\cite{zhang2025tora, wang2025levitor, yin2023dragnuwa, wu2024draganything, shi2024motion, li2025image, namekata2024sgi2v}. Quantitative comparison and qualitative comparison results are shown below.

\vspace{-0.5em}
\paragraph{Quantitative comparison.}
To compare MagicMotion with previous works, we use the first 49 frames of each video from DAVIS and MagicBench as the ground truth video. Since some methods~\cite{namekata2024sgi2v, wang2025levitor, yin2023dragnuwa, wu2024draganything, shi2024motion, li2025image} do not support video generation up to 49 frames in length, we uniformly sample N frames from these 49 frames for evaluation, where N represents the video length that each method support.
We leverage the mask and box annotations from these selected frames as trajectory inputs for mask or box-based methods. 
The center point of each frame's mask is extracted as input for point or flow-based methods~\cite{zhang2025tora, wang2025levitor, yin2023dragnuwa, wu2024draganything, shi2024motion, li2025image}. 

As shown in Table~\ref{tab:comparison}, our method outperforms all previous approaches across all metrics both on MagicBench and DAVIS, demonstrating its ability to produce higher-quality videos and more precise trajectory control.
Additionally, we evaluate each method’s performance on MagicBench based on the number of controlled objects. As shown in Fig.~\ref{fig:comparisons_radar}, our method achieves the best results across all object number categories, demonstrating the superiority of our approach.

\vspace{-0.5cm}
\paragraph{Qualitative comparison.}
Qualitative comparison results are shown in Fig.~\ref{fig:comparisons}, with the input image, prompt and trajectory provided. 
As shown in Fig.~\ref{fig:comparisons}, Tora~\cite{zhang2025tora} accurately controls the motion trajectory but struggles to maintain the shape of the objects. While DragAnything~\cite{wu2024draganything}, ImageConductor~\cite{li2025image}, and MotionI2V~\cite{shi2024motion} have difficulty preserving the consistency of the original subject, resulting in substantial deformation in subsequent frames. Meanwhile, DragNUWA~\cite{yin2023dragnuwa}, LeviTor~\cite{wang2025levitor}, and SG-I2V~\cite{namekata2024sgi2v} frequently produce artifacts and inconsistencies in fine details. In contrast, MagicMotion allows moving objects to follow the specified trajectory smoothly while preserving high video quality.

\subsection{Ablation Studies}
In this section, we present ablation studies to validate the effectiveness of our MagicData dataset. Additionally, we demonstrate how our progressive training procedure and latent segment loss enhance the model’s understanding of precise object shapes under sparse control conditions, thereby improving trajectory control accuracy.

\vspace{0.2cm}
\noindent\textbf{Ablations on Dataset.}
~~To verify the effectiveness of MagicData, we constructed an ablation dataset by combining two public VOS datasets, MeViS~\cite{MeViS} and MOSE~\cite{MOSE}.
For a fair comparison, we trained MagicMotion stage2 for one epoch using either MagicData or the ablation dataset as the training set, both initialized with the same stage1 weights. We then evaluated the models on both MagicBench and DAVIS.

\begin{figure}[ht]
    \centering
    \vspace{-1.4em}
    \includegraphics[width=\linewidth]{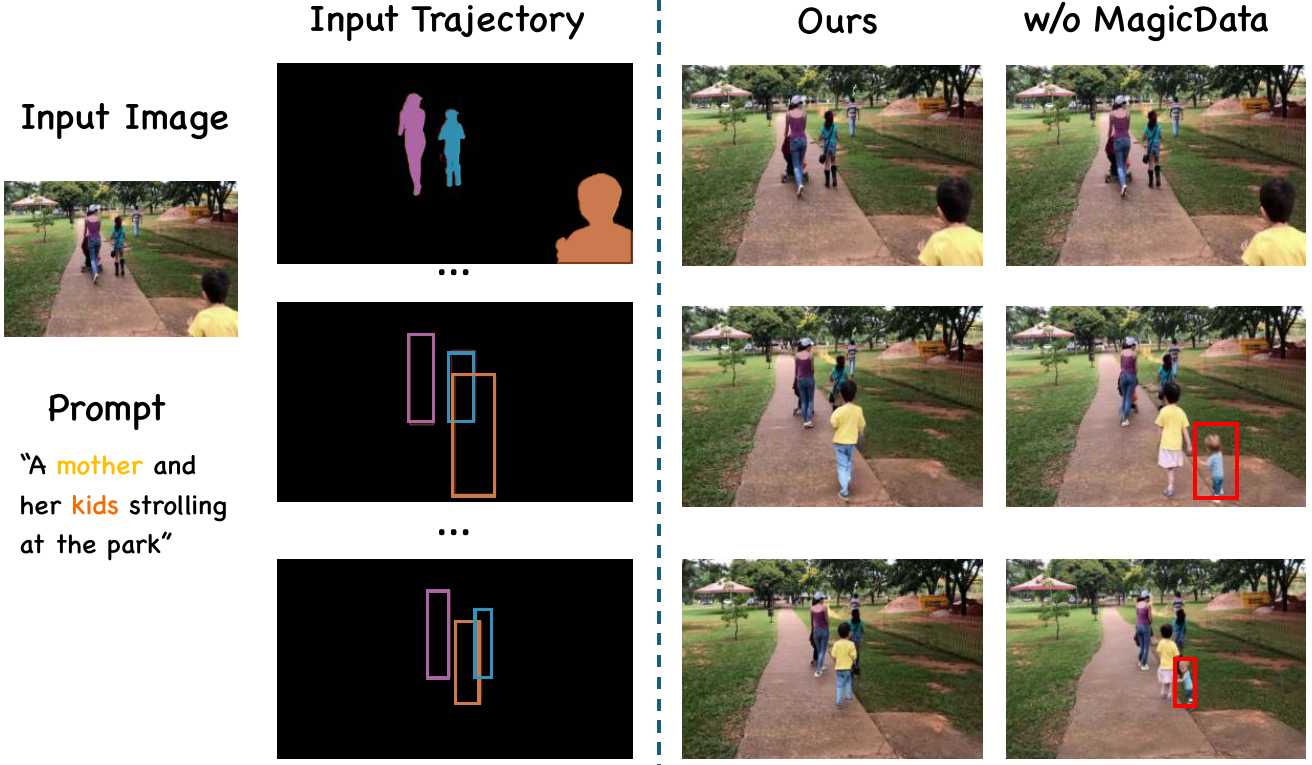}
    \caption{Ablation Study on MagicData. Not using MagicData causes the model to generate an unexpected child.}
    \label{fig:ablation dataset}
    \vspace{-1.4em}
\end{figure}

As shown in Table~\ref{tab:ablation dataset}, the model trained on MagicData outperforms the one trained on the ablation dataset across all metrics. Qualitative comparisons are shown in Fig.~\ref{fig:ablation dataset}. In this case, we aim to gradually move the boy in the lower right corner to the center of the image. However, not using MagicData results to an unexpected child appears next to the boy.
In contrast, the model trained with MagicData performs well, moving the boy along the specified trajectory while maintaining video quality.

\begin{table}[h]
\resizebox{\linewidth}{!}{
\begin{tabular}{@{}l|llll@{}}
\toprule
\multirow{2}{*}{\textbf{Method}} & \multicolumn{4}{c}{\textbf{MagicBench/DAVIS}}                                                 \\ \cmidrule(l){2-5} 
                                 & FID($\downarrow$)     & FVD($\downarrow$)      & M\_IoU\%($\uparrow$) & B\_IoU\%($\uparrow$) \\ \midrule
w/o MagicData                    & 24.24/50.91          & 125.19/768.27          & 73.39/50.95             & 78.66/70.05            \\ \midrule
Ours                             & \textbf{15.17/50.35} & \textbf{107.21/760.95} & \textbf{76.61/53.94}    & \textbf{81.45/72.84}   \\ \bottomrule
\end{tabular}
}
\caption{Ablation Study on MagicData. The model trained with MagicData outperforms the one trained without it across all metrics.}
\label{tab:ablation dataset}
\end{table}

\vspace{0.2cm}
\noindent\textbf{Ablations on Progressive Training Procedure.}
~~Progressive Training Procedure allows the model to leverage the weights learned in the previous stage, incorporating dense trajectory control information when trained with sparse trajectory conditions.
To validate the effectiveness of this approach, we train the model from scratch for one epoch using bounding boxes as trajectory conditions. We then compare its performance with MagicMotion stage2.

\begin{figure}[ht]
    \centering
    \vspace{-0.8em}
    \includegraphics[width=\linewidth]{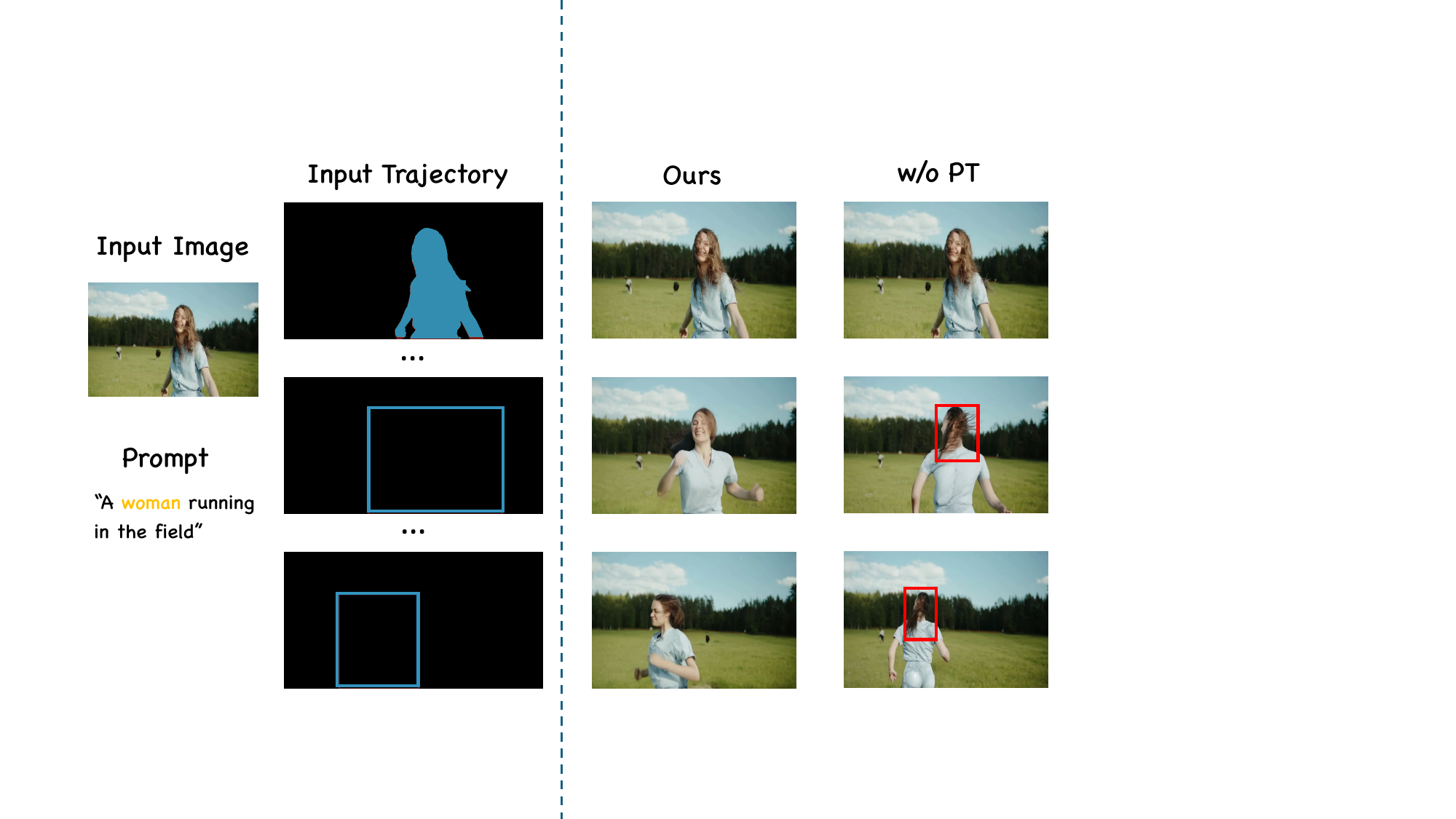}
    \caption{Ablation study on the progressive training procedure. Without it, the generated head shapes become noticeably distorted.}
    \label{fig:ablation training}
    \vspace{-1.4em}
\end{figure}

As shown in Table~\ref{tab:ablation progressive & latent}, excluding Progressive Training Procedure weakens the model’s ability to perceive object shapes, ultimately reducing the accuracy of trajectory control. Qualitative comparisons in Fig.~\ref{fig:ablation training} further illustrate these effects, where the model trained without Progressive Training Procedure turns the woman’s head entirely into hair.

\begin{table}[ht]
\resizebox{\linewidth}{!}{
\begin{tabular}{@{}l|ll|ll@{}}
\toprule
\multirow{2}{*}{\textbf{Method}} & \multicolumn{2}{c|}{\textbf{MagicBench}}                                              & \multicolumn{2}{c}{\textbf{DAVIS}}                                                   \\ \cmidrule(l){2-5} 
                                 & \multicolumn{1}{c}{M\_IoU\%($\uparrow$)} & \multicolumn{1}{c|}{B\_IoU\%($\uparrow$)} & \multicolumn{1}{c}{M\_IoU\%($\uparrow$)} & \multicolumn{1}{c}{B\_IoU\%($\uparrow$)} \\ \midrule
w/o LSL                          & 74.65                                      & 78.02                                      & 49.19                                      & 64.19                                     \\
w/o PT                           & 74.61                                      & 79.49                                      & 49.22                                      & 67.34                                     \\ \midrule
Ours                             & \textbf{76.61}                             & \textbf{81.45}                             & \textbf{53.94}                             & \textbf{72.84}                            \\ \bottomrule
\end{tabular}
}
\caption{Ablation Study on Progressive Training (PT) and Latent Segment Loss (LSL). Experimental results demonstrate that these techniques enhance the model with better comprehension on fine-grained object shapes.}
\label{tab:ablation progressive & latent}
\end{table}

\noindent\textbf{Ablations on Latent Segment Loss.}
~~Latent Segment Loss makes the model predict dense segmentation masks while training with sparse trajectories, enhancing its ability to perceive fine-grained object shapes under sparse conditions. To evaluate the effectiveness of this technique, we train the model from stage1 for one epoch using bounding boxes as trajectory conditions and compare its performance with MagicMotion stage2.
Table~\ref{tab:ablation progressive & latent} shows that the absence of Latent Segment Loss reduces the model’s ability on object shapes, leading to less precise trajectory control.  

%% file: sec/5_conclusion.tex
\section{Conclusion}
In this paper, we proposed MagicMotion, a trajectory-controlled image-to-video generation method that uses a ControlNet-like architecture to integrate trajectory information into the diffusion transformer. We employed a progressive training strategy, allowing MagicMotion to support three levels of trajectory control: dense masks, bounding boxes and sparse boxes. We also utilized Latent Segment Loss to enhance the ability of the model to perceive fine-grained object shapes when only provided with sparse trajectory conditions.
Additionally, we presented MagicData, a high-quality annotated dataset for trajectory-controlled video generation, created through a robust data pipeline.
Finally, we introduced MagicBench, a large-scale benchmark for evaluating trajectory-controlled video generation. MagicBench not only assessed video quality and trajectory accuracy but also took the number of controlled objects into account.
Extensive experiments on both MagicBench and DAVIS demonstrated the superiority of MagicMotion compared to previous works.
\\
\textbf{Acknowledge} This work was supported in part by the National Natural Science Foundation of China (Grant 62032006 and Grant 62472098).

%% file: sec/X_supplementary.tex
\clearpage
\appendix
\setcounter{page}{1}
\maketitle

\section{Ablations on Latent Segment Loss}
\begin{figure*}[ht]
    \centering
    \vspace{-0.8em}
    \includegraphics[width=\linewidth]{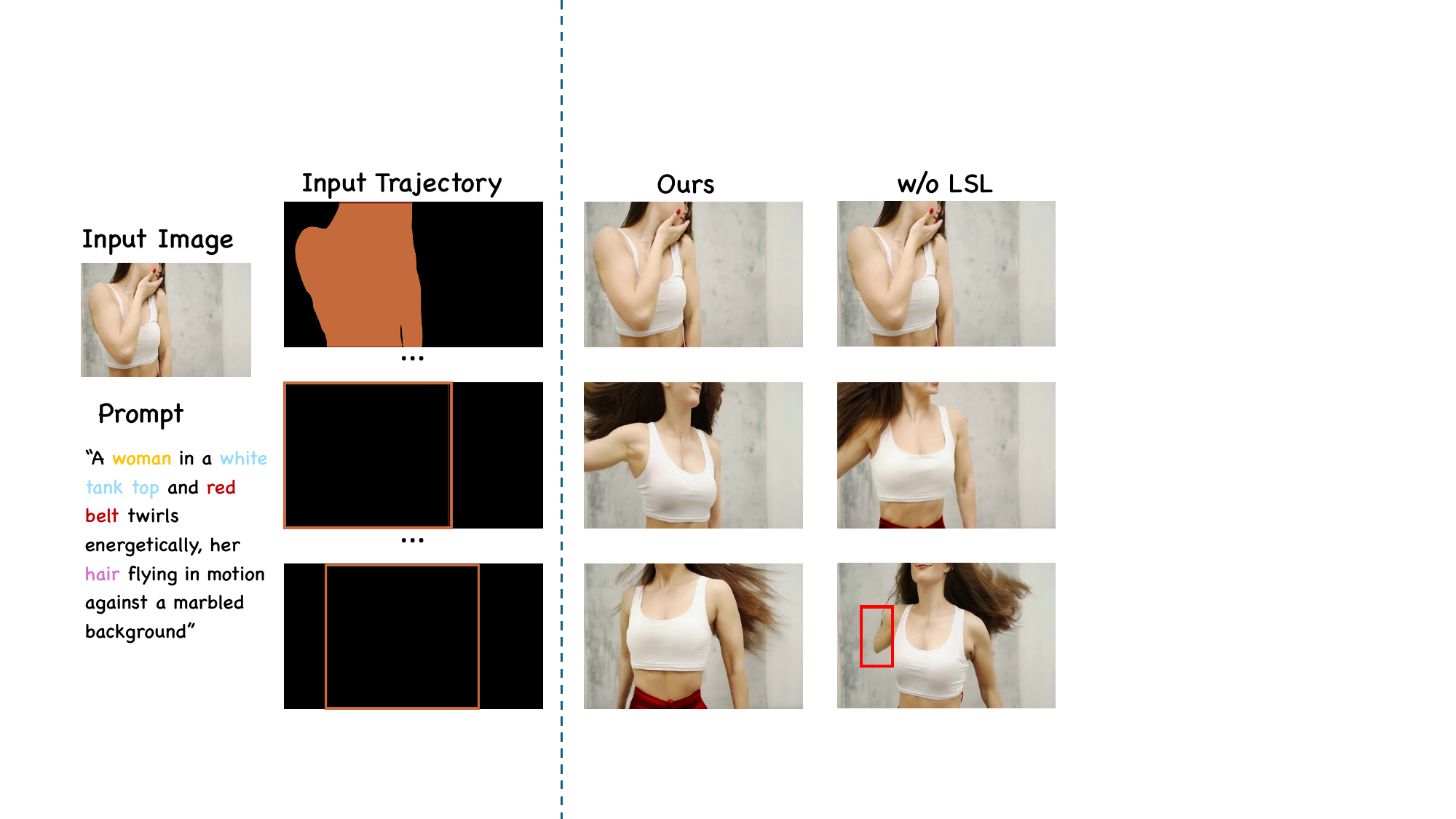}
    \caption{Ablation Study on latent segment loss. Without it, the generated arms appear partially missing.}
    \label{fig:ablation segment}
    \vspace{-1.4em}
\end{figure*}

We present qualitative results from the ablation study on Latent Segment Loss. As shown in Fig.~\ref{fig:ablation segment}, removing the Latent Segment Loss significantly reduces the model’s ability to capture object shapes, resulting in less accurate trajectory control. For instance, without this loss, the woman’s arm in the generated video appears incomplete.

\begin{figure*}[ht]
    \centering
    \includegraphics[width=\linewidth]{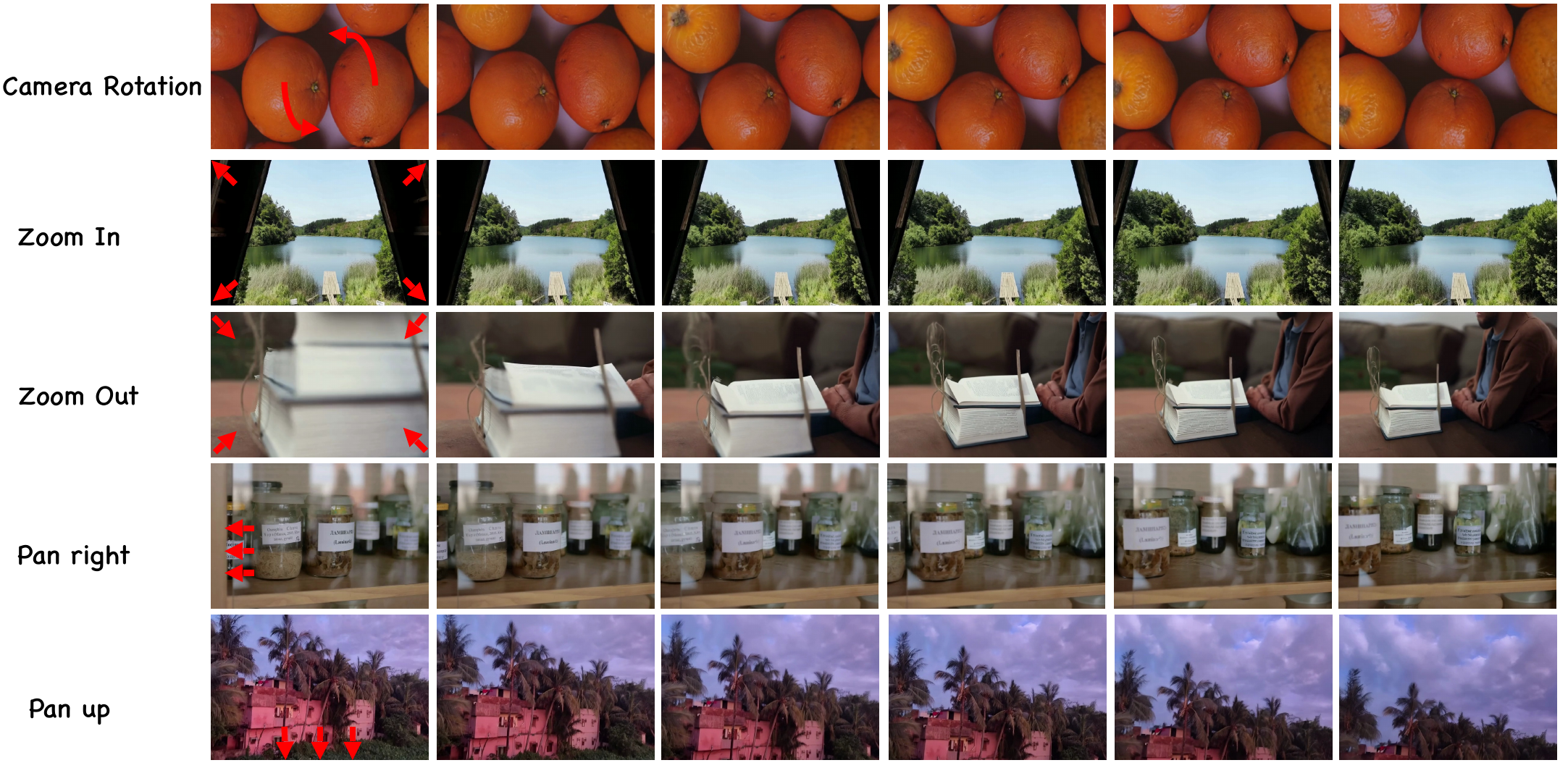}
    \caption{Camera motion controlled results. By setting specific trajectory conditions, MagicMotion can control camera movements.}
    \label{fig:supp camera motion}
\end{figure*}
\section{Additional Experiments}
We conducted additional experiments using MagicMotion under various task settings, including camera motion control and video editing. We also generate videos by applying different motion trajectories to a single input image.

\paragraph{Camera Motion Control.}
As shown in Fig.~\ref{fig:supp camera motion}, MagicMotion enables precise control over camera motion, allowing for operations such as rotation, zoom, and pan.
In the first row of Fig.~\ref{fig:supp camera motion}, we enclose oranges within bounding boxes and apply rotation to the boxes. This results in a video with a simulated camera rotation effect.
In the second and third rows, we adjust the size of the foreground object’s bounding box to control its perceived distance from the camera, effectively achieving zoom-in and zoom-out effects.
In the last two rows, we shift the bounding box to the left and downward, creating the effect of the camera moving in the opposite direction.

\begin{figure*}[h]
    \centering
    \includegraphics[width=\linewidth]{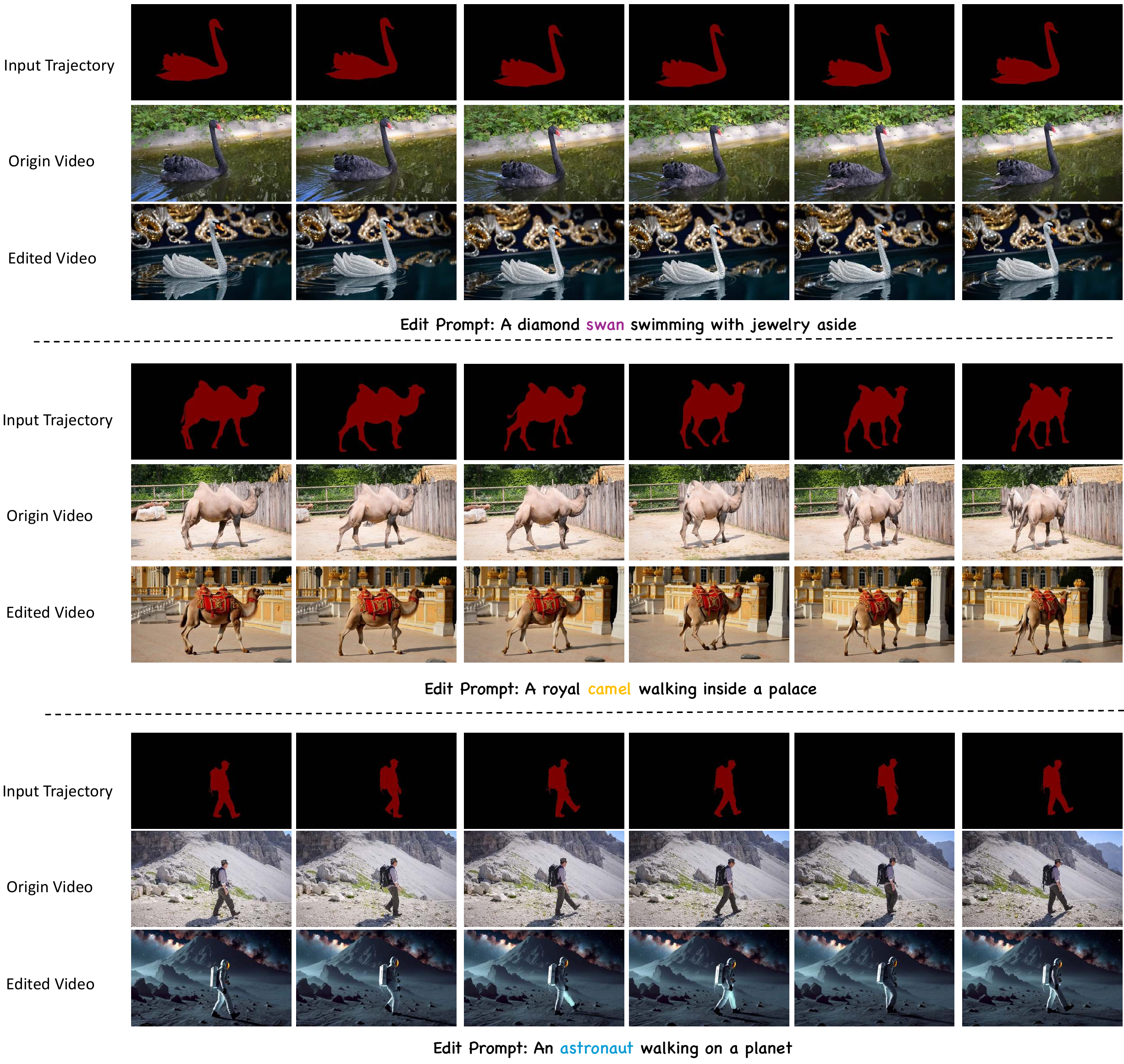}
    \caption{Video Editing Results. We use FLUX~\cite{flux2024} to edit the first-frame image and MagicMotion Stage1 to move the foreground objects following the trajectory of the origin video.}
    \label{fig:supp editing}
\end{figure*}

\paragraph{Video Editing.}
As shown in Fig.~\ref{fig:supp editing}, MagicMotion can be applied to video editing to generate high-quality videos. Specifically, we first use FLUX~\cite{flux2024} to edit the first frame of the original video, which serves as the input for MagicMotion. Then, we extract the segment mask of the original video and use it as trajectory guidance for the MagicMotion Stage1. 
Using this approach, we transform a black swan into a diamond swan, make the camel walk in a majestic palace, and turn a hiking backpacker into an astronaut.

\begin{figure*}[h]
    \centering
    \includegraphics[width=\linewidth]{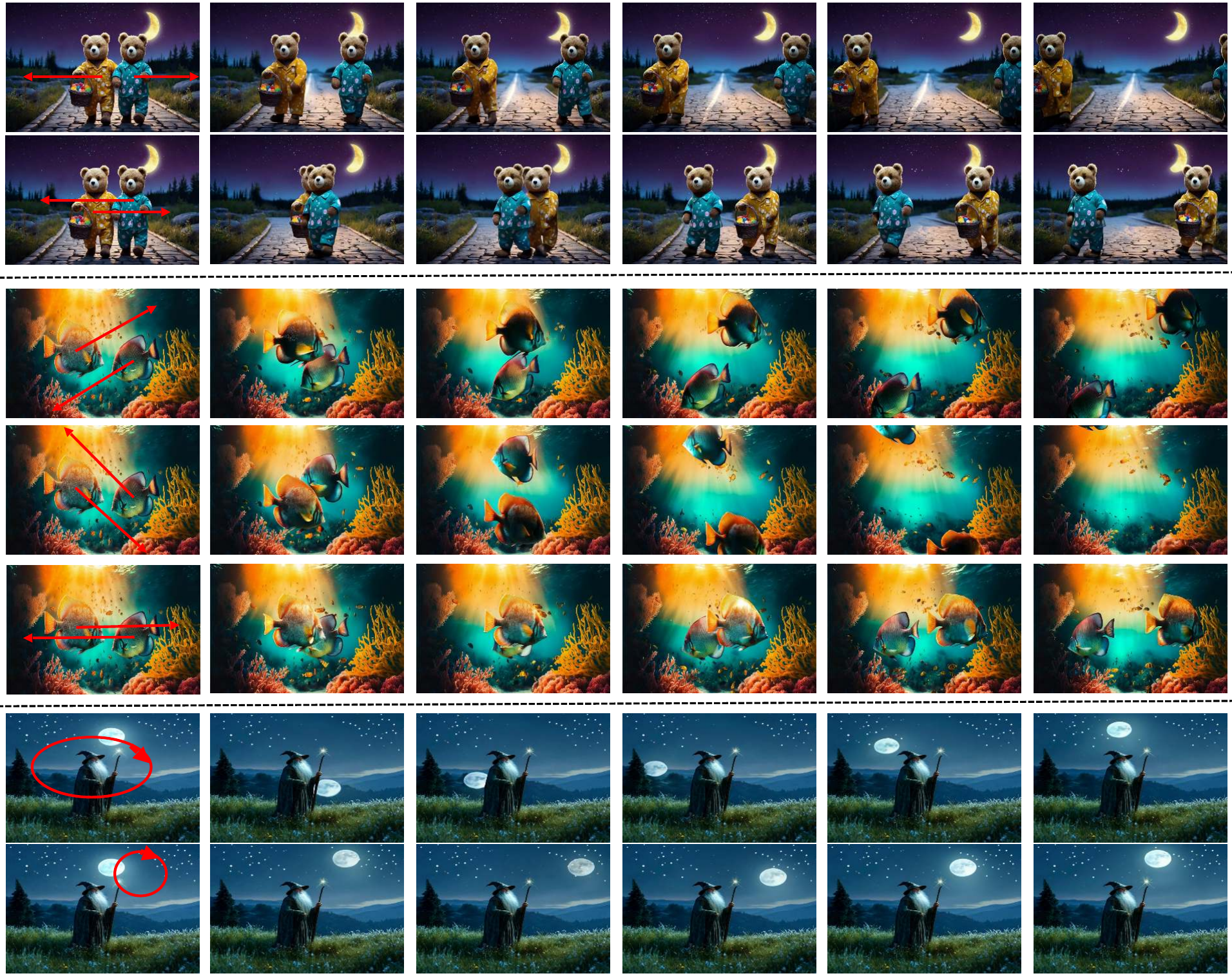}
    \caption{MagicMotion can generate videos using the same input image and different trajectories (marked by red arrows).}
    \label{fig:supp same image different traj}
\end{figure*}

\paragraph{Same input image with different trajectories}
Extensive experiments have demonstrated that MagicMotion enables objects to move along specified trajectories, generating high-quality videos. To further showcase the capabilities of MagicMotion, we use stage2 of MagicMotion to animate objects from the same input image along different motion trajectories.
As shown in Fig.~\ref{fig:supp same image different traj}, MagicMotion successfully animates two bears, two fish, and the moon, each following their designated paths.

\section{Latent Segment Masks}
\begin{figure*}[h]
    \centering
    \includegraphics[width=\linewidth]{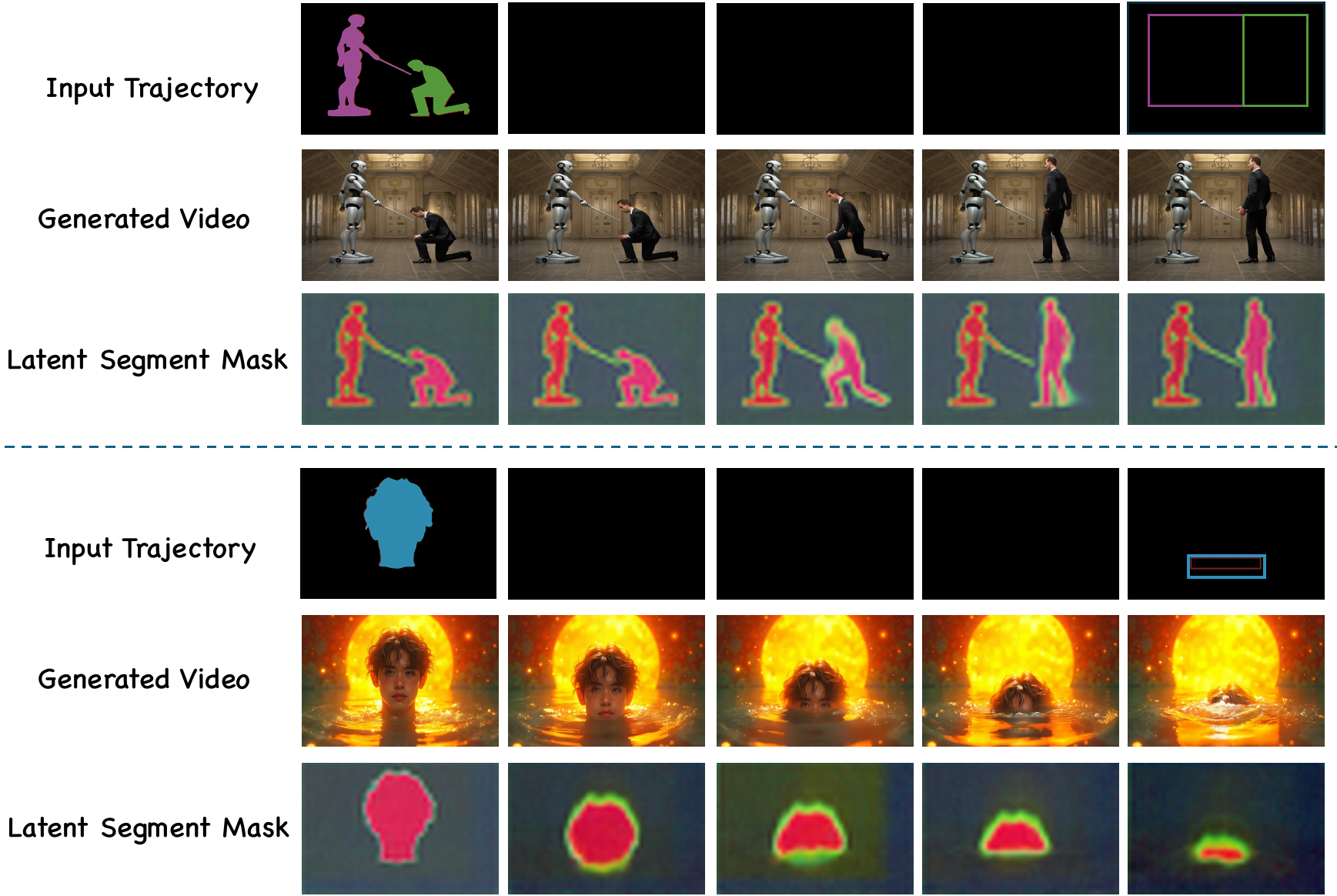}
    \caption{Latent Segment Masks visualization. MagicMotion can predict out latent segment masks of each frame even when only provided with sparse bounding boxes guidance.}
    \label{fig:supp latent segment}
\end{figure*}

In this section, we provide a more detailed demonstration of Latent Segmentation Masks. Specifically, we use MagicMotion Stage 3 to predict the latent segmentation masks for each frame based on sparse bounding box conditions.
As shown in Fig.~\ref{fig:supp latent segment}, MagicMotion accurately predicts the Latent Segmentation Masks throughout dynamic scenes, such as a man gradually standing up to face a robot and a boy’s head slowly sinking into the water. This holds true for frames where only the bounding box trajectory is available and even for frames where no trajectory information is provided at all.

\section{Additional Comparison results}
\paragraph{Baseline Comparisons.}
As shown in Table~\ref{tab: supp baseline comparison}, we provide a comparison of the backbones used by each method, along with the supported video generation length and resolution. 

\begin{table}[h]
\centering
\resizebox{\linewidth}{!}{
\begin{tabular}{l|ccc}
\hline
                                                                                      & Resolution & Length & Base Model                                                  \\ \hline
Motion-I2V~\cite{shi2024motion}                                 & 320*512    & 16     & AnimateDiff~\cite{guo2023animatediff} \\
ImageConductor~\cite{li2025image}      & 256*384    & 16     & AnimateDiff~\cite{guo2023animatediff} \\ \hline
DragAnything~\cite{wu2024draganything}                          & 320*576    & 25     & SVD~\cite{blattmann2023stable}        \\
LeViTor~\cite{wang2025levitor}                                  & 288*512    & 16     & SVD~\cite{blattmann2023stable}        \\
DragNUWA~\cite{yin2023dragnuwa}                                 & 320*576    & 14     & SVD~\cite{blattmann2023stable}        \\
SG-I2V~\cite{namekata2024sgi2v}                                 & 576*1024   & 14     & SVD~\cite{blattmann2023stable}        \\ \hline
Tora~\cite{zhang2025tora} & 480*720    & 49     & CogVideoX~\cite{yang2024cogvideox}    \\
\textbf{MagicMotion-CogVideoX}                                                                  & 480*720    & 49     & CogVideoX~\cite{yang2024cogvideox}    \\ \hline
\textbf{MagicMotion-Wan1.3B}                                                          & 480*832    & 81     & Wan~\cite{wan2025}                 \\ \hline
\end{tabular}
}
\caption{Comparisons on each method's backbone.}
\label{tab: supp baseline comparison}
\end{table}

\paragraph{Quantitative Comparisons on different object number}
Due to space constraints, we only included radar charts in the main text to compare the performance of different methods in controlling varying numbers of objects on MagicBench. Here, we provide the specific quantitative results. As shown in Table~\ref{tab:magicbench objnum1&2}, Table~\ref{tab:magicbench objnum3&4}, and Table~\ref{tab:magicbench objnum5&6}, MagicMotion consistently outperforms other methods across all metrics by a significant margin, especially when the number of moving objects is large. This demonstrates that other methods exhibit poorer performance when controlling a larger number of objects.

\begin{table*}[h]
\centering
\resizebox{\linewidth}{!}{
\begin{tabular}{c|cccc|cccc}
\hline
\multirow{2}{*}{\textbf{Method}}                                                                       & \multicolumn{4}{c|}{\textbf{MagicBench (Object Number = 1)}}                                                                                                          & \multicolumn{4}{c}{\textbf{MagicBench (Object Number = 2)}}                                                                                                                   \\ \cline{2-9} 
                                                                                                       & \multicolumn{1}{c}{FID($\downarrow$)} & \multicolumn{1}{c}{FVD($\downarrow$)} & \multicolumn{1}{c}{Mask\_IoU($\uparrow$)} & \multicolumn{1}{c|}{Box\_IoU($\uparrow$)} & \multicolumn{1}{c}{FID($\downarrow$)} & \multicolumn{1}{c}{FVD($\downarrow$)} & \multicolumn{1}{c}{Mask\_IoU($\uparrow$)} & \multicolumn{1}{c}{Box\_IoU($\uparrow$)} \\ \hline
Motion-I2V~\cite{shi2024motion}                                 & 78.3245                             & 660.8655                              & 0.6057                                    & 0.7142                                    & 90.4269                              & 867.6057                              & 0.5078                                    & 0.5938                                   \\
ImageConductor~\cite{li2025image}      & 106.1569                              & 674.4987                              & 0.5706                                    & 0.6974                                    & 109.2959                            & 879.8890                              & 0.4786                                    & 0.5579                                   \\
DragAnything~\cite{wu2024draganything}                          & 72.2494                              & 884.6453                              & 0.6706                                    & 0.8088                                    & 85.6511                              & 940.4941                              & 0.6148                                    & 0.7232                                   \\
LeViTor~\cite{wang2025levitor}                                  & 80.9714                              & 492.4725                              & 0.5536                                    & 0.7057                                    & 94.0106                              & 542.8533                              & 0.4352                                    & 0.5364                                   \\
DragNUWA~\cite{yin2023dragnuwa}                                 & 76.3963                              & 610.0368                              & 0.6699                                    & 0.7769                                    & 81.2921                              & 624.3985                              & 0.6033                                    & 0.6809                                   \\
SG-I2V~\cite{namekata2024sgi2v}                                 & 60.8617                              & 547.8107                              & 0.7144                                    & 0.8668                                    & 71.8699                              & 619.2791                              & 0.6315                                    & 0.7378                                   \\
Tora~\cite{zhang2025tora} & 60.2737                              & 805.0145                              & 0.6468                                    & 0.7776                                    & 73.6499                              & 795.8535                              & 0.5509                                    & 0.6584                                   \\ \hline
\textbf{Ours (Stage1-Wan1.3B)}                                                                 & 48.4211                          & \underline{461.1011}                          & \underline{0.8518}                          & \underline{0.9186}                          & \underline{49.3071}                          & \underline{438.2138}                          & \underline{0.8143}                          & 0.8161                          \\
\textbf{Ours (Stage2-Wan1.3B)}                                                                 & 51.2964                          & 555.1390                          & 0.6693                          & 0.8163                          & 55.5581                          & 569.0549                          & 0.6318                          & 0.7454                          \\ \hline
\textbf{Ours (Stage1-CogVideoX)}                                                                                  & \textbf{42.3523}                               & \textbf{473.2179}                              & \textbf{0.9359}                                    & \textbf{0.9607}                                    & \textbf{48.6525}                              & \textbf{428.3430}                              & \textbf{0.9080}                                    & \textbf{0.9097}                                   \\
\textbf{Ours (Stage2-CogVideoX)}                                                                                  & \underline{46.1080}                              & 564.1036                              & 0.7363                                    & 0.9017                                    & 52.7296                              & 550.5857                              & 0.6931                                   & \underline{0.8256}                                   \\ \hline
\end{tabular}
}
\caption{Quantitative Comparison results on MagicBench with moving objects number equals to 1 / 2.}
\label{tab:magicbench objnum1&2}
\end{table*}

\begin{table*}[h]
\centering
\resizebox{\linewidth}{!}{
\begin{tabular}{c|cccc|cccc}
\hline
\multirow{2}{*}{\textbf{Method}}                                                                       & \multicolumn{4}{c|}{\textbf{MagicBench (Object Number = 3)}}                                                                                                          & \multicolumn{4}{c}{\textbf{MagicBench (Object Number = 4)}}                                                                                                                   \\ \cline{2-9} 
                                                                                                       & \multicolumn{1}{c}{FID($\downarrow$)} & \multicolumn{1}{c}{FVD($\downarrow$)} & \multicolumn{1}{c}{Mask\_IoU($\uparrow$)} & \multicolumn{1}{c|}{Box\_IoU($\uparrow$)} & \multicolumn{1}{c}{FID($\downarrow$)} & \multicolumn{1}{c}{FVD($\downarrow$)} & \multicolumn{1}{c}{Mask\_IoU($\uparrow$)} & \multicolumn{1}{c}{Box\_IoU($\uparrow$)} \\ \hline
Motion-I2V~\cite{shi2024motion}                                 & 89.3195                              & 842.6530                              & 0.5366                                    & 0.6076                                    & 83.2024                              & 744.5470                              & 0.6018                                    & 0.6484                                   \\
ImageConductor~\cite{li2025image}      & 114.7662                              & 927.3884                              & 0.5169                                    & 0.5578                                    & 110.8037                              & 832.4498                              & 0.5679                                    & 0.5417                                   \\
DragAnything~\cite{wu2024draganything}                          & 85.1437                              & 925.2795                              & 0.6332                                    & 0.6625                                    & 71.8865                              & 901.7427                              & 0.6946                                    & 0.7148                                   \\
LeViTor~\cite{wang2025levitor}                                  & 93.3111                              & 607.7522                              & 0.3809                                    & 0.4671                                    & 90.4561                              & 688.8164                              & 0.3555                                    & 0.4044                                   \\
DragNUWA~\cite{yin2023dragnuwa}                                 & 79.3941                              & 642.4184                              & 0.6420                                    & 0.7012                                    & 71.0587                              & 512.5130                              & 0.7085                                    & 0.7250                                   \\
SG-I2V~\cite{namekata2024sgi2v}                                 & 70.0600                              & 520.2733                              & 0.6531                                    & 0.7068                                    & 56.9318                               & 460.1303                              & 0.7145                                    & 0.7423                                   \\
Tora~\cite{zhang2025tora} & 66.6571                              & 742.4080                              & 0.5926                                    & 0.6356                                    & 64.0669                              & 779.0798                              & 0.6226                                    & 0.6312                                   \\ \hline
\textbf{Ours (Stage1-Wan1.3B)}                                                        & 51.7821                          & 458.4122                          & \underline{0.8236}                          & \underline{0.8122}                          & 47.8972                          & 463.4390                          & \underline{0.8598}                          & \underline{0.8288}                          \\
\textbf{Ours (Stage2-Wan1.3B)}                                                        & 56.9454                          & 568.4777                          & 0.6632                          & 0.7089                          & 50.0364                          & 548.9445                          & 0.7112                          & 0.7252                          \\ \hline
\textbf{Ours (Stage1)}                                                                                                   & \textbf{42.9636}                               & \textbf{421.0036}                              & \textbf{0.9107}                                    & \textbf{0.8797}                                    & \textbf{37.6524}                               & \textbf{396.4754}                              & \textbf{0.9231}                                    & \textbf{0.8896}                                   \\
\textbf{Ours (Stage2)}                                                                                                   & \underline{45.4721}                               & \underline{440.2373}                              & 0.7562                                    & 0.8097                                    & \underline{37.4172}                               & \underline{442.0640}                              & 0.7998                                    & 0.8253                                   \\ \hline
\end{tabular}
}
\caption{Quantitative Comparison results on MagicBench with moving objects number equals to 3 / 4.}
\label{tab:magicbench objnum3&4}
\end{table*}

\begin{table*}[h]
\centering
\resizebox{\linewidth}{!}{
\begin{tabular}{c|cccc|cccc}
\hline
\multirow{2}{*}{\textbf{Method}}                                                                       & \multicolumn{4}{c|}{\textbf{MagicBench (Object Number = 5)}}                                                                                                          & \multicolumn{4}{c}{\textbf{MagicBench (Object Number \textgreater 5)}}                                                                                                                   \\ \cline{2-9} 
                                                                                                       & \multicolumn{1}{c}{FID($\downarrow$)} & \multicolumn{1}{c}{FVD($\downarrow$)} & \multicolumn{1}{c}{Mask\_IoU($\uparrow$)} & \multicolumn{1}{c|}{Box\_IoU($\uparrow$)} & \multicolumn{1}{c}{FID($\downarrow$)} & \multicolumn{1}{c}{FVD($\downarrow$)} & \multicolumn{1}{c}{Mask\_IoU($\uparrow$)} & \multicolumn{1}{c}{Box\_IoU($\uparrow$)} \\ \hline
Motion-I2V~\cite{shi2024motion}                                 & 84.2231                              & 582.0029                              & 0.6295                                    & 0.6267                                    & 79.9077                              & 923.2557                              & 0.4899                                    & 0.4511                                   \\
ImageConductor~\cite{li2025image}      & 117.8054                              & 857.3489                              & 0.5180                                    & 0.4737                                    & 106.3168                              & 963.6862                              & 0.4536                                    & 0.3442                                   \\
DragAnything~\cite{wu2024draganything}                          & 79.7015                              & 710.5812                              & 0.7050                                    & 0.7011                                    & 87.1509                              & 719.2442                              & 0.6534                                    & 0.6045                                   \\
LeViTor~\cite{wang2025levitor}                                  & 90.8332                              & 578.5567                              & 0.3913                                    & 0.4281                                    & 92.5265                              & 763.3157                              & 0.2812                                    & 0.2768                                   \\
DragNUWA~\cite{yin2023dragnuwa}                                 & 77.2094                              & 435.9205                              & 0.7253                                    & 0.6988                                    & 78.5956                              & 549.7680                              & 0.6638                                    & 0.5709                                   \\
SG-I2V~\cite{namekata2024sgi2v}                                 & 65.2560                               & 453.1147                              & 0.7431                                    & 0.7367                                    & 95.7569                              & 596.4075                              & 0.6616                                    & 0.6211                                   \\
Tora~\cite{zhang2025tora} & 67.3827                              & 709.1618                              & 0.6228                                    & 0.6111                                    & 77.2571                              & 907.9254                              & 0.4976                                    & 0.4866                                   \\ \hline
\textbf{Ours (Stage1-Wan1.3B)}                                                        & 40.8559                          & 384.3586                          & \underline{0.8645}                          & 0.8112                          & 39.6606                          & \textbf{395.2139}                          & \underline{0.8136}                          & 0.6726                          \\
\textbf{Ours (Stage2-Wan1.3B)}                                                        & 46.7040                          & 491.7079                          & 0.7461                          & 0.7210                          & 42.0005                          & 454.2931                          & 0.6941                          & 0.5861                          \\ \hline
\textbf{Ours (Stage1)}                                                                                                                    & \textbf{39.4636}                               & \underline{374.6467}                              & \textbf{0.9155}                                    & \textbf{0.8600}                                    & \underline{39.2044}                               & 449.3122                              & \textbf{0.9012}                                    & \textbf{0.7653}                                   \\
\textbf{Ours (Stage2)}                                                                                                                    & \underline{40.0656}                               & \textbf{350.5010}                              & 0.8106                                    & \underline{0.8123}                                    & \textbf{37.1917}                               & \underline{396.0661}                              & 0.8004                                    & \underline{0.7124}                                   \\ \hline
\end{tabular}
}
\caption{Quantitative Comparison results on MagicBench with moving objects number equals to 5 / above 5.}
\label{tab:magicbench objnum5&6}
\end{table*}

\begin{figure*}[h]
    \centering
    \includegraphics[width=\linewidth]{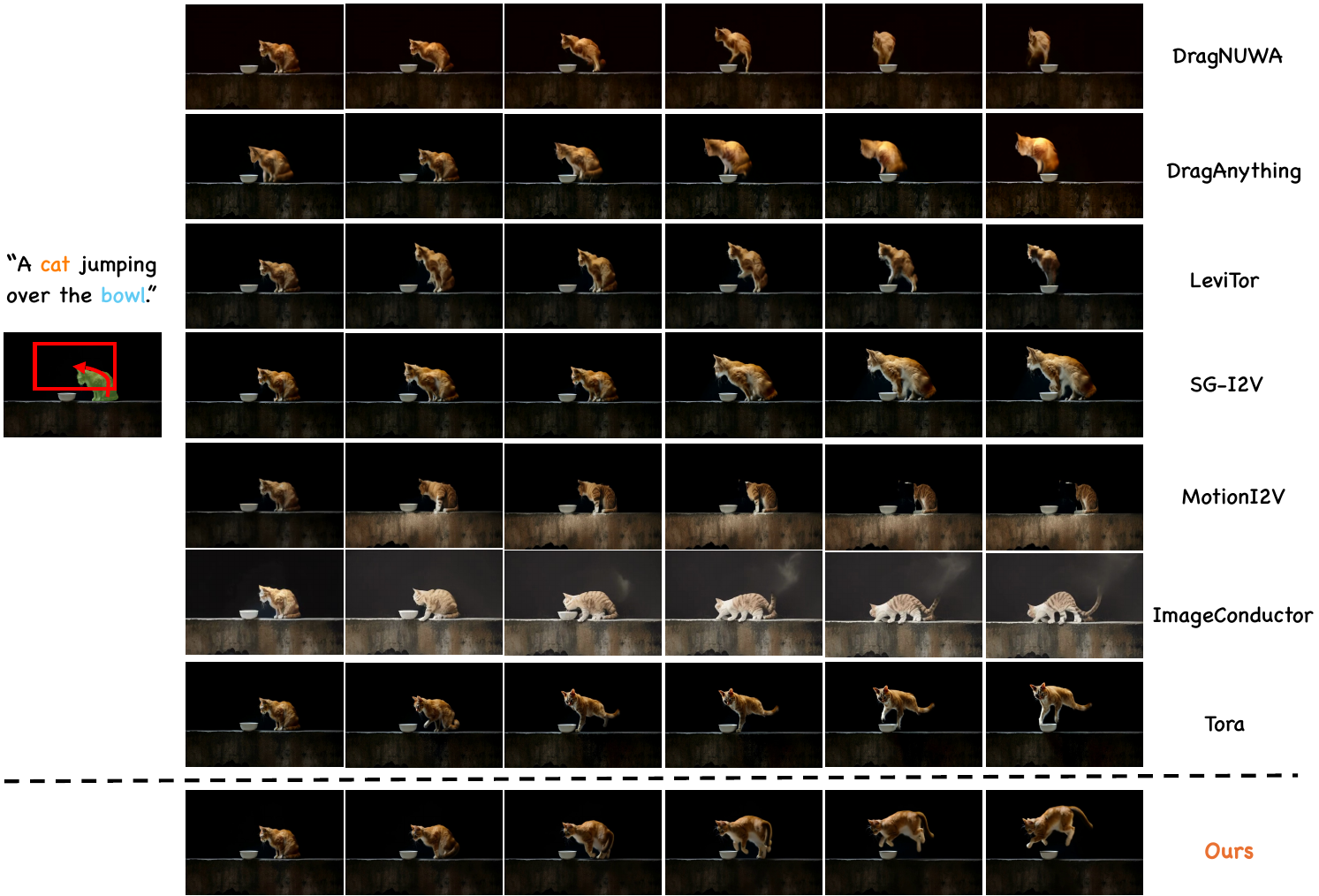}
    \caption{Qualitative Comparisons Results. MagicMotion successfully control the cat jumping over the bowl, while all other methods exhibit significant defects.}
    \label{fig:supp comparisons1}
\end{figure*}

\begin{figure*}[h]
    \centering
    \includegraphics[width=\linewidth]{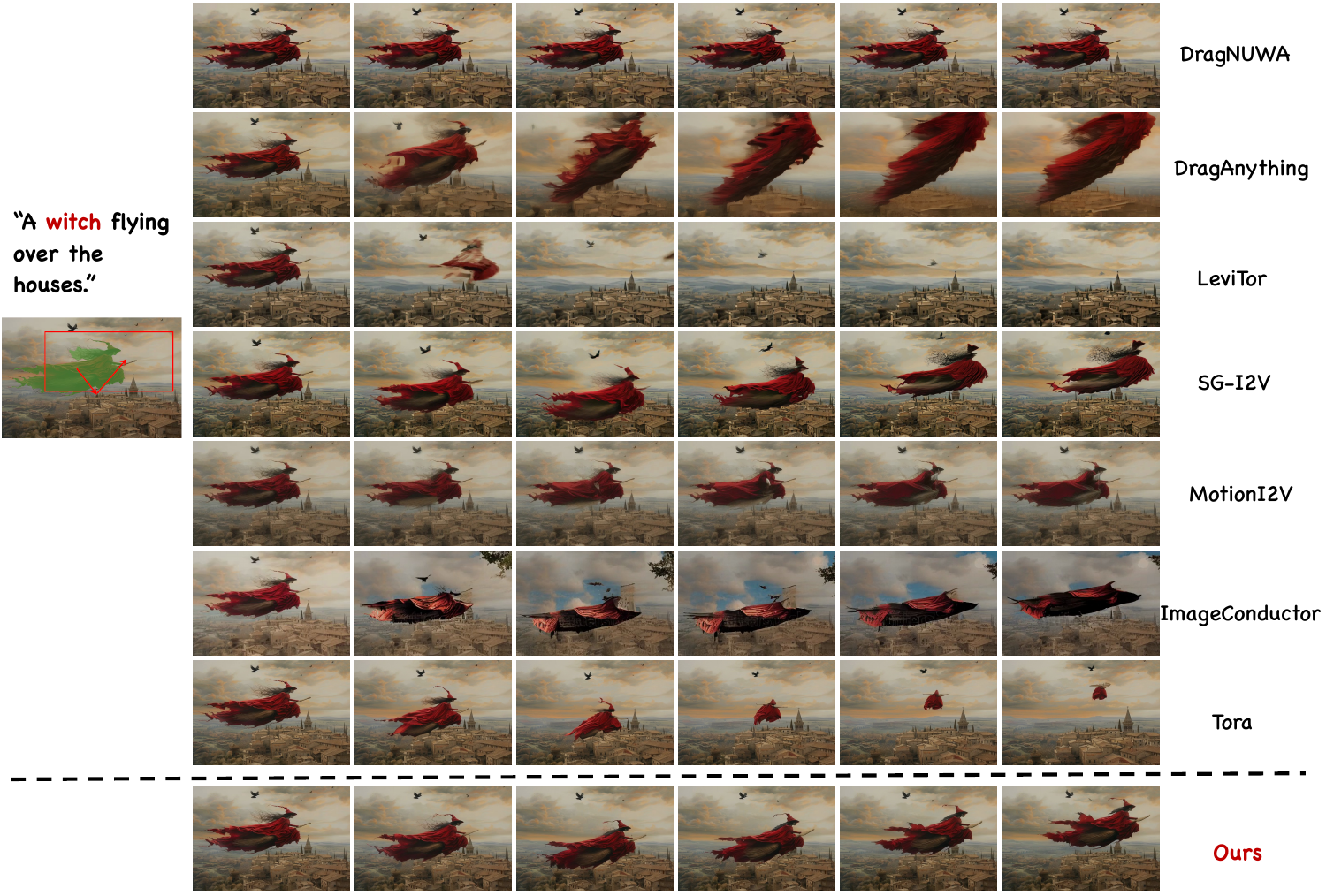}
    \caption{Qualitative Comparisons Results. MagicMotion successfully control the witch flying over the input trajectory, while all other methods exhibit significant defects.}
    \label{fig:supp comparisons2}
\end{figure*}

\begin{figure*}[h]
    \centering
    \includegraphics[width=\linewidth]{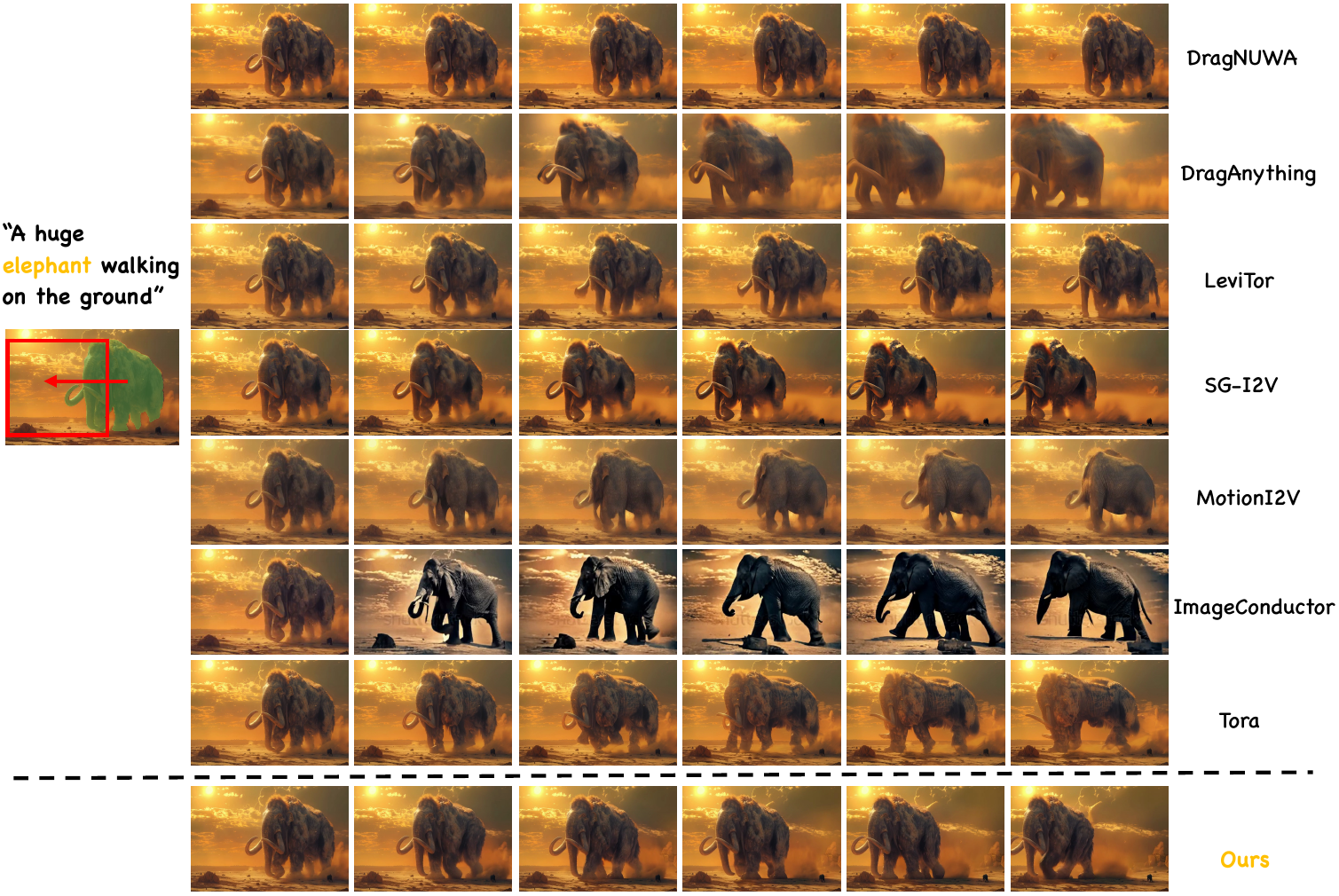}
    \caption{Qualitative Comparisons Results. MagicMotion successfully control the elephant walking along the input trajectory, while all other methods exhibit significant defects.}
    \label{fig:supp comparisons3}
\end{figure*}

\begin{figure*}[h]
    \centering
    \includegraphics[width=\linewidth]{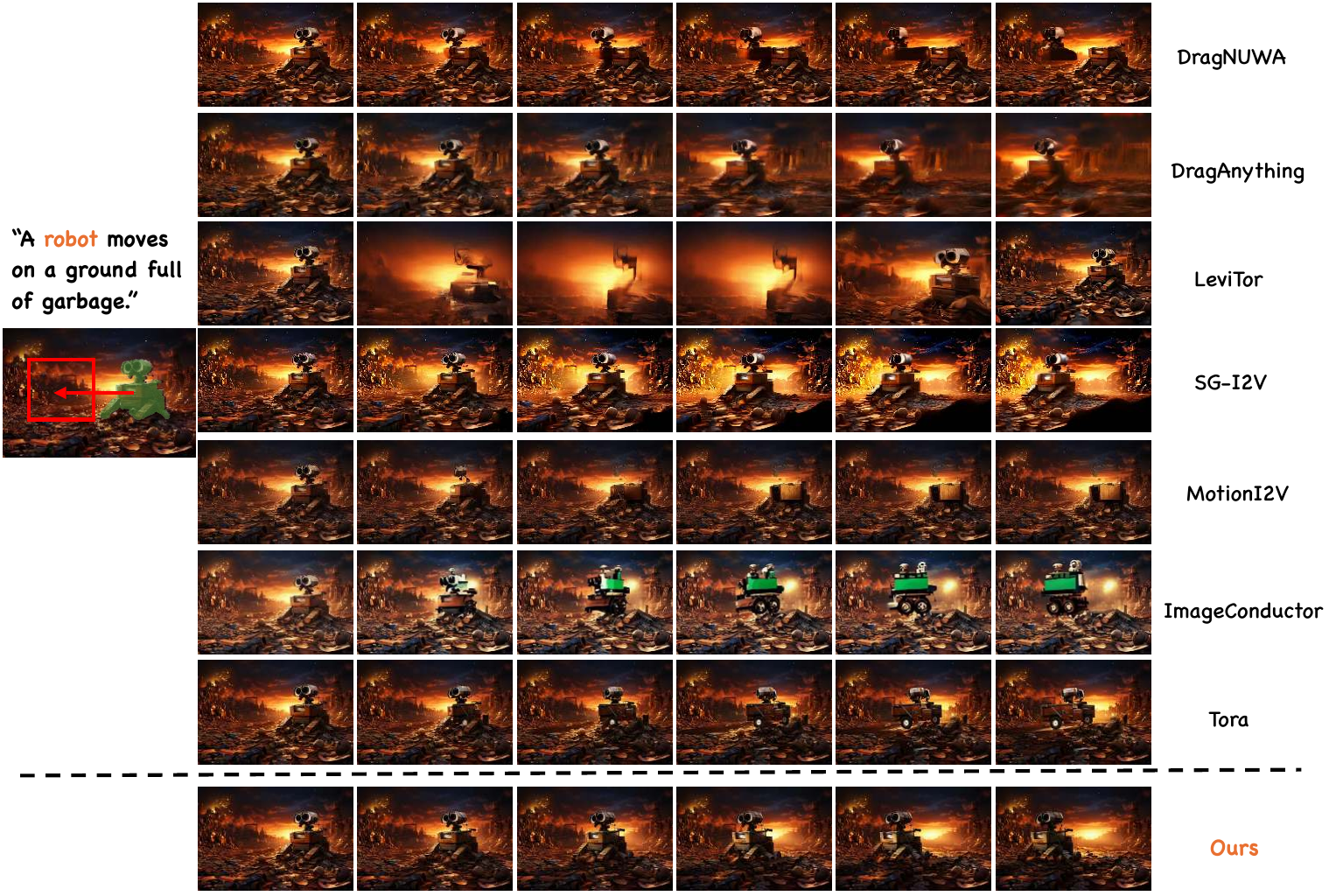}
    \caption{Qualitative Comparisons Results. MagicMotion successfully control the robot moving along the input trajectory, while all other methods exhibit significant defects.}
    \label{fig:supp comparisons4}
\end{figure*}

\begin{figure*}[h]
    \centering
    \includegraphics[width=\linewidth]{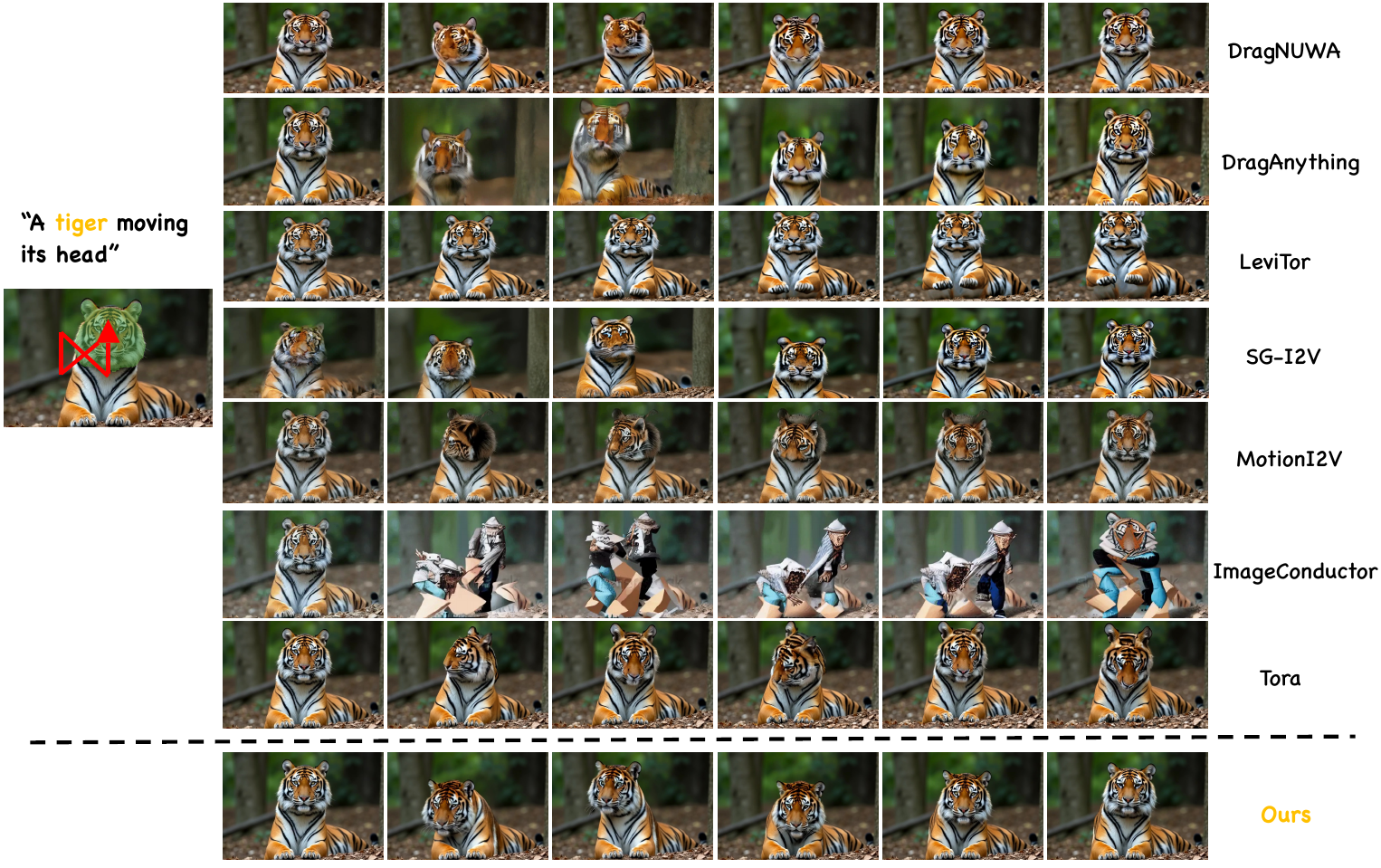}
    \caption{Qualitative Comparisons Results. MagicMotion successfully control the tiger's head moving along the input trajectory, while all other methods exhibit significant defects.}
    \label{fig:supp comparisons5}
\end{figure*}

\paragraph{More qualitative comparison results.}
In this section, we provide additional qualitative comparison results with previous works.
As shown in Fig.\ref{fig:supp comparisons1}, Fig.\ref{fig:supp comparisons2}, Fig.\ref{fig:supp comparisons3}, Fig.\ref{fig:supp comparisons4}, and Fig.~\ref{fig:supp comparisons5}, MagicMotion accurately controls object trajectories and generates high-quality videos, while other methods exhibit significant defects. For fully rendered videos, we refer the reader to “$\texttt{Supplementary video.mp4}$” in supplementary material.

\section{Additional Ablation Results.}
\begin{figure*}[h]
    \centering
    \includegraphics[width=\linewidth]{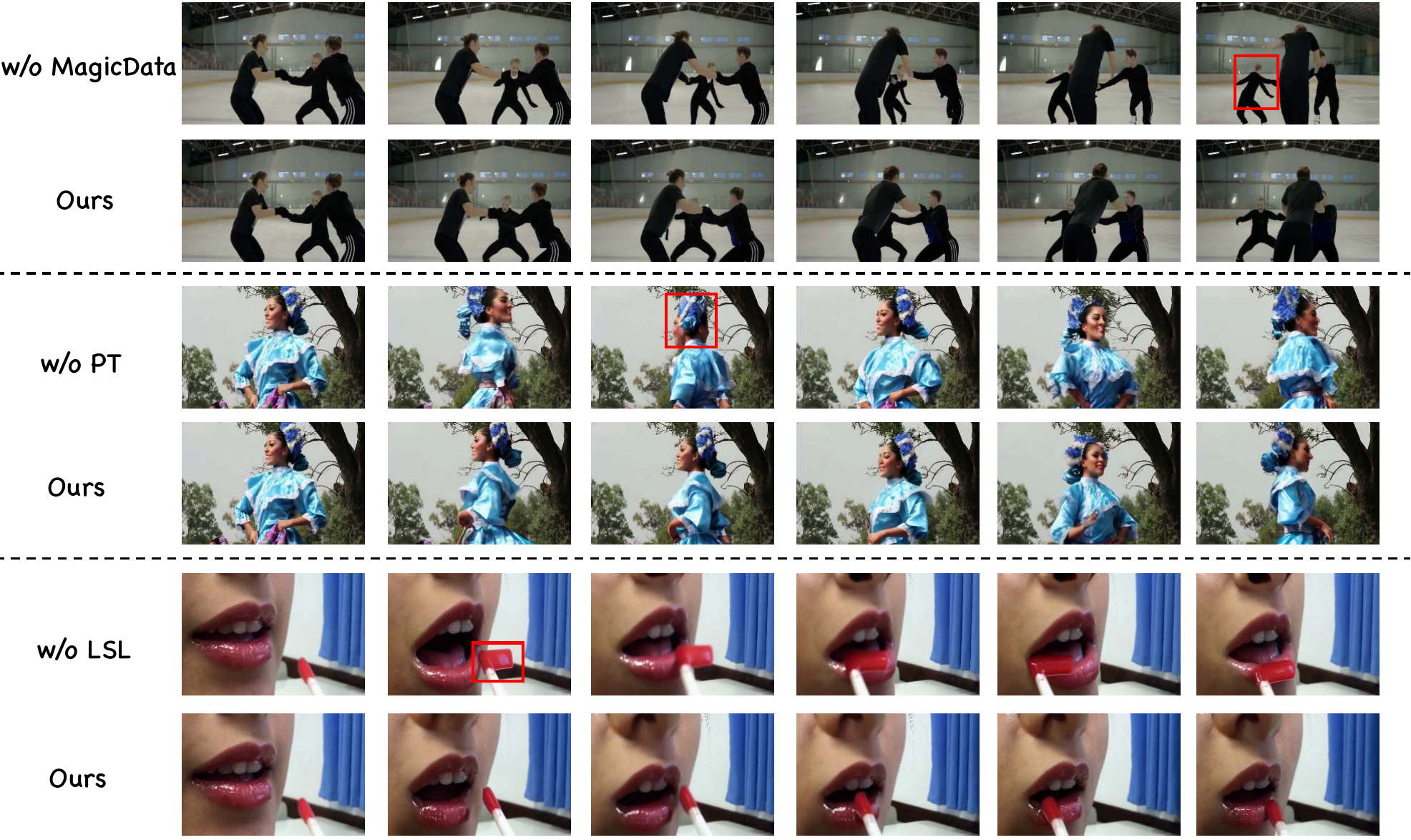}
    \caption{Additional Ablation results.}
    \label{fig:additional ablation}
\end{figure*}

Here, we provide additional qualitative comparison results from the ablation study.
As shown in Fig.~\ref{fig:additional ablation}, not using MagicData for training results in the generation of a woman with an extra hand. Not using the Progressive Training Procedure results in significant defects, such as a dancing woman showing severe issues when turning, with a second face appearing where her hair should be. Additionally, without the Latent Segment Loss, the woman’s lipstick is distorted into a rectangular shape.

\section{More Details on MagicData}
\begin{figure*}[h]
    \centering
    \includegraphics[width=\linewidth]{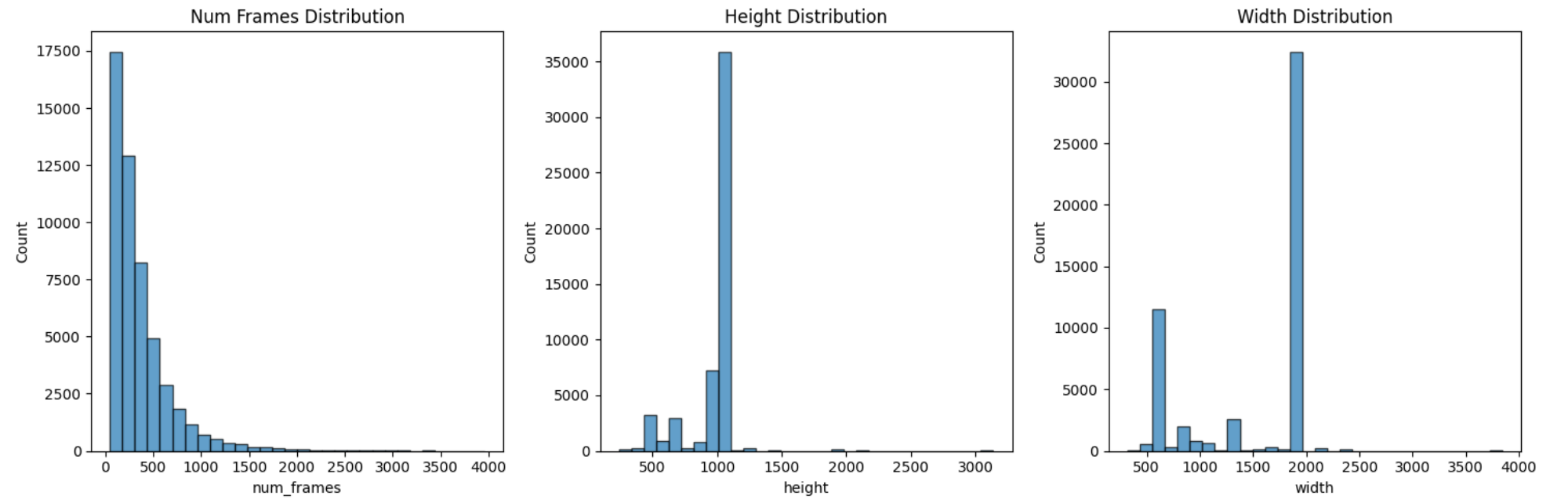}
    \caption{Detail information on MagicData.}
    \label{fig:magicdata details}
\end{figure*}

Here, we provide some detailed statistical information about MagicData. On average, each video in MagicData contains 346 frames, with a typical height of 999 pixels and a width of 1503 pixels.
For a more comprehensive understanding of the distribution and variability across the dataset, please refer to Fig.~\ref{fig:magicdata details}, which visualizes the detailed distribution of video frame counts, heights, and widths. During training, these videos are resized to 48 frames and converted to a 720p resolution.

\section{More Details on MagicBench}
For evaluation purposes, all videos in MagicBench are sampled to 49 frames and resized to a resolution of 720p. MagicBench is categorized into 6 classes based on the number of annotated foreground objects. Below, we provide one video example for each category, offering a more intuitive understanding of MagicBench.

\begin{figure*}[h]
    \centering
    \includegraphics[width=\linewidth]{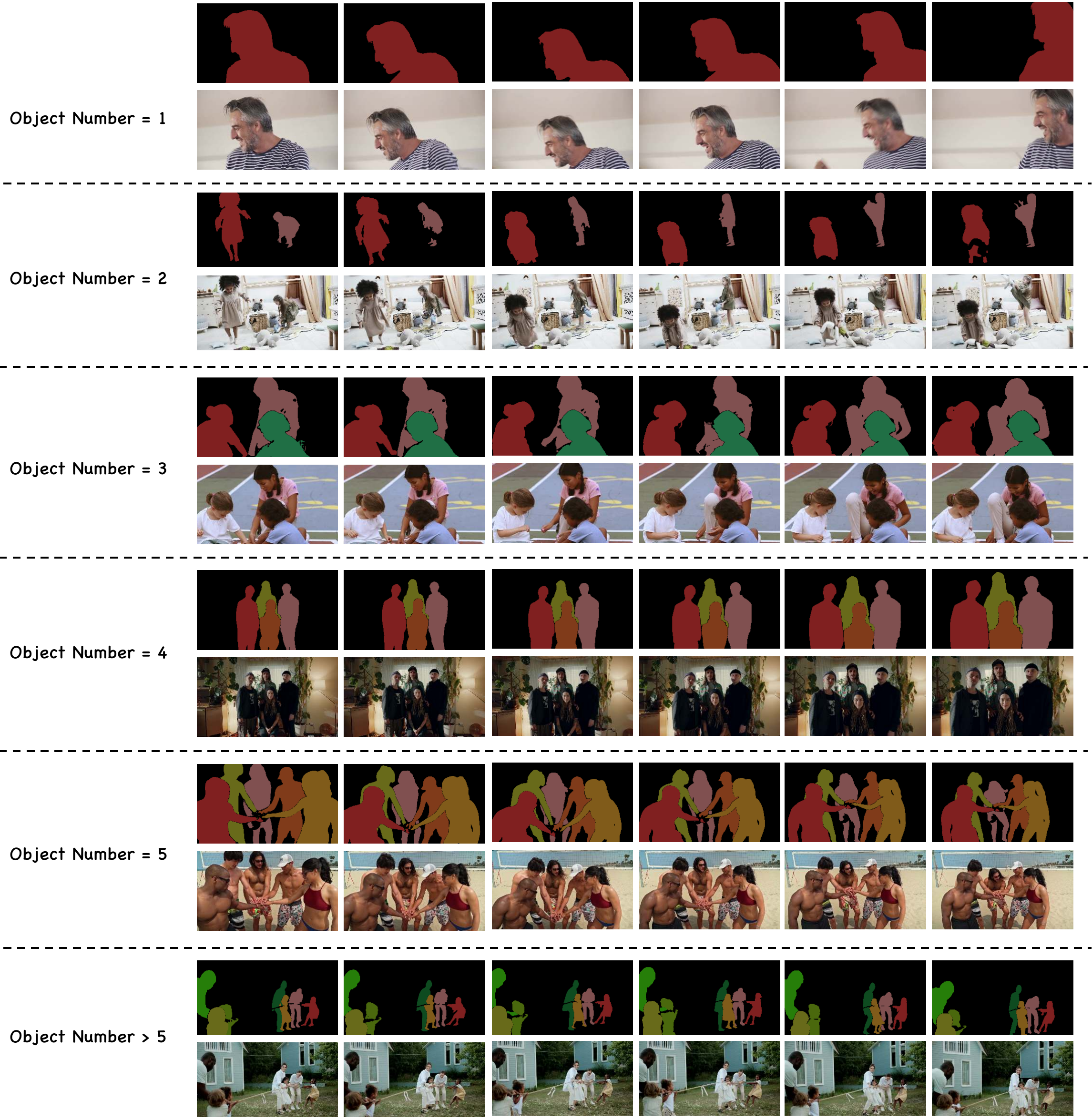}
    \caption{MagicBench visualization. We provide one video as a visual example for each object number category..}
    \label{fig:magicbench visualization}
\end{figure*}